\UseRawInputEncoding
\documentclass[12pt]{article}
\usepackage{amsfonts}
\usepackage{mathrsfs}
\usepackage{amsthm}
\usepackage{multirow}
\usepackage{bbding}
\usepackage{amssymb}
\usepackage{amsmath}
\usepackage{graphicx,color}
\usepackage{rotating}
\usepackage{bm}
\usepackage{epstopdf}
\usepackage{cite}

\newcommand{\bea}{\begin{eqnarray}}
\newcommand{\eea}{\end{eqnarray}}
\newcommand{\Bea}{\begin{eqnarray*}}
\newcommand{\Eea}{\end{eqnarray*}}
\newcommand{\ba}{\begin{array}}
\newcommand{\ea}{\end{array}}
\newcommand{\bt}{\begin{tabular}}
\newcommand{\et}{\end{tabular}}
\newcommand{\btb}{\begin{table}}
\newcommand{\etb}{\end{table}}
\newcommand{\bc}{\begin{center}}
\newcommand{\ec}{\end{center}}
\newcommand{\beq}{\begin{equation}}
\newcommand{\eeq}{\end{equation}}

\makeatletter

\newcommand{\Rmnum}[1]{\expandafter\@slowromancap\romannumeral #1@}
\makeatother

\begin{document}

\title{A Correlation-Ratio Transfer Learning and Variational Stein's Paradox
}
\author{Lu Lin$^1$, Weiyu Li$^1$\footnote{The corresponding
author. Email: liweiyu@sdu.edu.cn. The research was
supported by NNSF projects (11971265) of China.} 
\\
\small $^1$Zhongtai Securities Institute for Financial Studies, Shandong University, Jinan, China\\
 }
\date{}
\maketitle

\vspace{-0.7cm}
\begin{abstract} \baselineskip=17pt
A basic condition for efficient transfer learning is the similarity between a target model and source models.
In practice, however, the similarity condition is difficult to meet or is even violated. Instead of the similarity condition, a bran-new strategy, linear correlation-ratio, is introduced in this paper to build an accurate relationship between the models. Such a correlation-ratio can be easily estimated by historical data or a part of sample. Then, a correlation-ratio transfer learning likelihood is established based on the correlation-ratio combination. On the practical side, the new framework is applied to some application scenarios, especially the areas of data streams and medical studies. Methodologically, some techniques are suggested for transferring the information from simple source models to a relatively complex target model. Theoretically, some favorable properties, including the global convergence rate, are achieved, even for the case where the source models are not similar to the target model. All in all, it can be seen from the theories and experimental results that the inference on the target model is significantly improved by the information from similar or dissimilar source models. In other words, a variational Stein's paradox is illustrated in the context of transfer learning.


\end{abstract}
\baselineskip=20pt

\setcounter{equation}{0}
\section{Introduction}
\subsection{Problem setup and article frame}

The models with heterogeneous distributions are frequently encountered in big data territory, especially in the environment of big data streams. The relevant heterogeneity has attracted much attention in the areas of statistics and machine learning. Transfer learning is a special methodology in machine learning to deal with a type of heterogeneity: there is a target model (or distribution), about which we wish to make inference, and there are some related but distinct source models (or distributions), from which most of training data are collected. Under this framework, the heterogeneity means that there exist certain differences between the target model and the source models, and the goal of transfer learning is to use the data not only from the target model but also from the source models to make an improved target inference. Up to now, various transfer learning methods have been proposed in the  literature of statistics and machine learning; for survey papers
on transfer learning, see for example  Storkey (2009), Pan and Yang (2009), and Weiss et al. (2016). Moreover, the methods of transfer learning have been widely used in many applications, including
computer vision  (Tzeng et al., 2017; Gong et al., 2012), speech recognition (Huang et al., 2013), and genre classification  (Choi et al., 2017).

For convenience, we consider for the moment the case of only a source model. Let $P$ be the source model containing response variable $Y^P$ and covariate $X^P$, and $Q$ be the target model with response variable $Y^Q$ and covariate $W^Q$. From the perspective of canonical transfer learning, the main reason why a method of transfer learning can transfer knowledge from $P$ to $Q$ for target inference is that the two models are supposed to be similar, or the level of heterogeneity between $P$ and $Q$ is relatively low. In practice, however, the similarity condition is difficult to meet or is even violated (the examples will be given later). We then are interested in the following issue:
\begin{itemize}\item {\it Develop an adaptive transfer learning strategy
capable of processing general source $P$ to improve
the inference on target $Q$ whether $P$ is similar to $Q$ or not.}
\end{itemize}
When the issue is first encountered,
it seems to be preposterous.
Actually, if model $P$ is independent of model $Q$, the phenomenon observed in this paper is a variational Stein's paradox (for the explanations see the Appendix, for the details of Stein's paradox see, e.g.,  Stein et al., 1956; James and Stein, 1992; Lindley, 1962; Efron and Morris, 1973 and
Stigler, 1990). In the area of transfer learning, the solution to this problem is interesting and challenging, but has not yet been explored so far, to the best of our knowledge.
\begin{itemize}
\item {\it  In this work, instead of using the similarity condition, a historical data condition of a part of variables is employed to establish a relationship between the models $P$ and $Q$. Based on the relationship, an accurate transfer learning likelihood is established. Consequently,
the inference on $Q$ can be significantly improved by the information from $P$ even if the two models are dissimilar. }\end{itemize}

This paper is then structured as follows. In the remainder of this section, some existing transfer learning methods are reviewed, the similarity conditions are discussed, and our main contributions are summarized. In Section 2, some motivating examples are discussed to build the relationship between target model and source models, and under the framework of generalized linear models (GLMs), a bran-new correlation strategy is proposed and then
a correlation-ratio transfer learning likelihood (cr-TLL) is established, without using any similarity condition. In Section 3, the adaptability of the cr-TLL to heterogeneous sources is demonstrated, theoretically, and the variational Stein's paradox is analyzed. In Section 4, the cr-TLL is extended into the case where the target model is a semiparametric model.
Simulation studies and real data analysis are presented in Section 5. Section 6 concludes the paper and discusses possible future work. The following additional materials are relegated to the Appendix: a brief description of Stein's paradox and the relationship to our model condition, the estimation methods for correlation-ratios, the iterative algorithm for some complex cases, the extension to non-linear models, some results of additional simulation studies,
the regularity conditions and the proofs of the main lemmas and theorems.

\subsection{Similarity conditions and related issues}
Before introducing our methodology, we first recall the familiar  similarity conditions and discuss the relevant topics on transfer learning. It is known that the similarity between source $P$ and target $Q$ is the most basic condition for efficient transfer learning, traditionally.
Up to now, several rules have been proposed in the transfer learning literature to measure the degree of similarity between $P$ and $Q$. The first method is to establish the bounds of divergence between $P$ and $Q$ to evaluate the level of similarity ( (Ben-David et al., 2006;
Blitzer et al., 2007; Mansour et al., 2009; Tian and Feng, 2021)).
Another line of work in the literature describes the level of similarity by measuring the covariate shift for the regime in which the conditional distributions of $Y^P$ and $Y^Q$ given $X^P$ and $W^Q$ are the same, but the marginal distributions of $X^P$ and $W^Q$ can be different (Shimodaira, 2000; Sugiyama
et al., 2007; Kpotufe and Martinet, 2021). On the other hand, the conditional distribution drift is a general framework and broadly arises in many practical problems (Pan and Yang, 2009; Weiss et al., 2016; Cai and Wei, 2021; Reeve et al., 2021). Under such a situation, the level of similarity is formulated in the existing literature by a transfer function $\phi(\cdot)$ together with some propensity constraints on $\phi(\cdot)$. Specifically, for some functions of interest $\eta_P(\cdot)$ and $\eta_Q(\cdot)$ (e.g., regression functions) under models $P$ and $Q$ respectively, the similarity relationship between the functions $\eta_P(\cdot)$ and $\eta_Q(\cdot)$ is formulated by $\eta_P(\cdot)=\phi(\eta_Q(\cdot))$ (see, e.g.,  Cai and Wei, 2021; Reeve et al., 2021).

In the area of parameter-transfer learning, the degree of similarity is evaluated in the following ways: the source model $P$ and target model $Q$ share some parameters or prior distributions of the hyperparameters of the models, or some parameters in $P$ and $Q$ are closed to each other (Lawrence and Platt, 2004; Evgeniou and Pontil, 2004; Schwaighofer et al., 2004; Bonilla et al., 2007; Gao et al., 2008; Xia et al., 2013; Tian and Feng, 2021; Li et al., 2020). Then, the transferred knowledge is encoded into the shared (or approximately equal) parameters or priors. By discovering them, the knowledge can be transferred across the tasks.

Without these similarity conditions, however, it is difficult by the existing methods to find useful knowledge from the source models. Moreover, in the procedures of transfer learning of these approaches, data-driven algorithms are often required to implement the transfer learning.

\subsection{The models under study and main contributions}

In this paper, we focus mainly on parameter estimation in parametric target model together with parametric or semiparametric multiple source models.
In our framework, all the parameters and variables in the source models may be entirely different from those in the target model, implying that the commonly used conditions for the similarity may be violated completely. In fact the models under study may be of both conditional distribution drift and covariate shift, and/or the source models may even be unrelated to the target model.
Such model variations (including conditional distribution drift, covariate shift and parametric heterogeneity) often appear in some application areas, especially in the framework of big data streams. For instance, U.S. carriers were not required to report
causes of flight delays to the Bureau of Transportation Statistics until June 2003, the related variables on the causes of flight delays were available only after June 2003. When parameter models are used for the risk analysis of airline delays, the covariates may change midway through the data streams (Rupp, 2007), implying that all the parameters in the two models may be entirely different (Wang et al., 2018; Lin et al., 2021). For more real-world examples on the issue, see Subsection 2.2, and the references, e.g.,  Certo (2003), Hood et al. (2004) and Desyllas and Sako (2013).

Therefore, it is desired to develop adaptive transfer learning together with efficient algorithm for the scenarios where source models and target model may have completely different parameters and variables, or the source models may be dissimilar to the target model in some sense.
Instead of canonical similarity conditions, it is supposed in this paper that some historical data or a part of sample of permanent variables are available before modeling. On the practical
side, the condition of historical information or a part of sample from a part of variables used in this paper is applied to some application scenarios, especially the areas
of data streams and medical studies (see the motivating examples in Subsection 2.2).
Based on this condition, a new transfer learning framework is introduced under GLMs, partially linear models and non-linear models.
Within the framework of exponential family distributions with both conditional distribution drifts and covariate shifts or heterogeneous parameters, some artificial  relationships among the parameters are formulated by a bran-new strategy: linear correlation-ratio matrix between the target model and source models. The correlation-ratio can be easily estimated by the historical data or a part of sample from permanent variables. It is worth pointing out that actually these are precise parameter relationships in the models, instead of the similarity conditions between the models. The formulas of describing the relationships are concise, even have explicit representations. Based on these accurate representations, the  correlation-ratio transfer learning likelihoods (cr-TLLs) are established that combine the target likelihood with the transferred likelihoods from the source models. The new methods are suitable for transferring the information from some simple source models to a relatively complex target model. Theoretically, the asymptotic properties of the transfer learning estimator are established and the efficiency is deeply investigated. An outstanding feature of the cr-TLL is what follows: the global convergence rate can be achieved, without any evident constraint on the similarity. Here the global convergence rate means that the transfer learning estimation is always $\sqrt n$-consistent whether the source models are related to the target model or not, where $n$ is the size of total samples from source models and target model. Thus, it is somewhat surprising that the new strategies can be applied to the case with related or unrelated sources. This implies that the phenomenon of a variational Stein's paradox is transparent (see the Appendix, and for more details see, e.g.,  Stein et al., 1956; Stigler, 1990). The theoretical properties and numerical results reveal the relationship between the artificial model combination (instead of the model similarity) and transfer learning efficiency, telling us when and how to use the transfer learning to improve the inference on the target model. Moreover, the algorithms are simple because the relationships between the parameters have simple representations.

Main contributions of this paper are
summarized as follows:
\begin{itemize} \item [(i)] Remarkably, an entirely new strategy of linear correlation-ratio is introduced to artificially combine the source models with target model. With the proposed combinations, the adaptive transfer learning method is applied to the target model with similar or dissimilar sources.
\item [(ii)] The relationships between the source models and target model have accurate expressions. Consequently, the transfer learning likelihood is accurate, containing complete information from sources.
    \item [(iii)] Theoretically and experimentally, the resulting transfer learning estimator is significantly improved by the information from similar or dissimilar sources, and particularly the global convergence rate in probability can be enhanced remarkably, especially for the case when the source models are somewhat similar to the target model.
        \item [(iv)] Methodologically, the transfer
learning is computationally tractable, without the need for any data-driven method to implement the inference procedure. \end{itemize}

\setcounter{equation}{0}
\section{cr-TLL for GLMs}

In the first four subsections, we introduce the cr-TLL for the case of a source model. In the last subsection, we consider the case with multiple source models.
The extension to generalized partially linear models and general non-linear models  will be given in Section 4 and the Appendix, respectively.

\subsection{Models}

\subsubsection{Data conditions.}
Denote by $P$ and $Q$ the two models with random variables $(X^P,Y^P)$ and $((X^Q,Z^Q),Y^Q)$ respectively, where response variables $Y^P\in\mathbb{S}_{Y^P}\subseteq \mathbb{R}$, $Y^Q\in \mathbb{S}_{Y^Q}\subseteq \mathbb{R}$, and covariates $X^P\in \mathbb{S}_{X^P}\subseteq\mathbb{R}^{d_1}$ and $(X^Q, Z^Q) \in \mathbb{S}_{X^Q}\times\mathbb{S}_{Z^Q}\subseteq\mathbb{R}^{d_1}\times \mathbb{R}^{d_2}$. In the framework of transfer learning, $P$ is thought of as a source model, from which most of training data are generated, and $Q$ is treated as a target model, from which some training data are collected, and about which we wish to make inference.
It is usually supposed that the collection of the data of some variables in model $Q$, for example $Z^Q$, is typically expensive or difficult, then the data volume of $Z^Q$ is relatively small.

In this paper, we further assume that some historical characteristics  (e.g., average values or expectations) or a part of sample of variables $(X^P,Y^P)$ and $(X^Q,Y^Q)$ are available before modeling.

Some motivating examples will be given in Subsection 2.2 to show the application backgrounds of the above historical or partial data condition.
On the other hand, the data of variable $Z^Q$ are available only after a time point or only in some special situation. We then call $(X^P,Y^P)$ and $(X^Q,Y^Q)$ permanent variables, and call $Z^Q$ emerging variable. The emergence of new variables is common due to, for example, negligence in data collection in the past, change of protocol, or advances in technology.
Also, we will present some motivating examples in Subsection 2.2 to further show the application backgrounds of the condition on permanent variables and emerging variable.

\subsubsection{GLMs and likelihood equations}

With the data conditions and characteristics aforementioned, our goal is to transfer the information from a simple source model $P$ to a complex target model $Q$ such that the inference on $Q$ is more accurate. Here we call $Q$ a complex model because it is designed to have a relatively complex structure: a GLM with more covariates under study in this section, or a semiparametric model that will be given in Section 4, or a general non-linear model that will be discussed in the Appendix.

We suppose the models follow exponential family distributions. It is known that exponential family is the most popular distribution pattern in statistics. This special form is chosen for mathematical convenience as well as for generality.

In this section, we choose the regime where $Y^P$ is distributed following a natural exponential family distribution. In this case,
the source model $P$ is supposed to be a GLM as
\begin{eqnarray}
E[Y^P|X^P]=g_P^{-1}(\bm\beta^TX^P),\end{eqnarray}
where $g_P^{-1}(\cdot)$ is a pre-specified inverse link function, and $\bm\beta$ is a $d_1$-dimensional vector of unknown parameters. Then, for a canonical GLM, the log-likelihood takes form: $$l_P(\bm\beta|Y^P,X^P)\propto Y^P\bm\beta^T X^P -G_P(\bm\beta^T X^P).$$ Here the exact form of $G_P(\cdot)$ depends on the model, for example, for linear regression, $G_P(u)$ is proportional to $u^2/2$, and for canonical (log-link) Poisson regression, $G_P(\cdot)=\exp(\cdot)$. Theoretically, the true value of $\bm\beta$ is defined as the solution to the following likelihood equation:
\begin{eqnarray}
E[Y^PX^P-X^P \dot{G}_P(\bm\beta^T X^P)]=0,
\end{eqnarray}
where $\dot{G}_P(\cdot)$ is the derivative of $G_P(\cdot)$.

Similarly, suppose that $Y^Q$ is distributed following an exponential family distribution. The main difference from the feature of $Y^P$ is that here $Y^Q$ depends on more covariates $X^Q$ and $Z^Q$. Then, the target model $Q$ is the following GLM:
\begin{eqnarray}\label{(Q-model)}E[Y^Q|X^Q,Z^Q]= g_Q^{-1}(\bm\gamma^TX^Q+
\bm\theta^T Z^Q),\end{eqnarray} where $g_Q^{-1}(\cdot)$ is an inverse link function as well, $\bm\gamma$ and $\bm\theta$ are respectively $d_1$- and $d_2$-dimensional vectors of unknown parameters.

The log-likelihood has the form:
$$l_Q(\bm\gamma,\bm\theta|Y^Q,X^Q,Z^Q)\propto Y^Q(\bm\gamma^TX^Q+
\bm\theta^T Z^Q) -G_Q(\bm\gamma^TX^Q+
\bm\theta^T Z^Q).$$
Also the exact form of $G_Q(\cdot)$ depends on the model. Theoretically, the true values of $\bm\gamma$ and $\bm\theta$ are defined as the solutions to the following joint likelihood equations:
\begin{eqnarray}
&& E[Y^QX^Q-X^Q \dot{G}_Q(\bm\gamma^TX^Q+
\bm\theta^T Z^Q)]=0,\\
&& E[Y^QZ^Q-Z^Q \dot{G}_Q(\bm\gamma^TX^Q+
\bm\theta^T Z^Q)]=0,
\end{eqnarray} where $\dot{G}_Q(\cdot)$ is the derivative of $G_Q(\cdot)$ as well.

In the parameter models, the goal of transfer learning is to make inference on the parameter vectors $\bm\gamma$ and $\bm\theta$ defined in likelihood equations (2.4) and (2.5) by the data from $Q$ together with information learned from source model $P$. It is worth pointing out that the above models do not belong to the existing frameworks of parameter-transfer learning because the parameter vector $\bm\beta$ in the source  model may be completely different from the parameter vectors $\bm\gamma$ and $\bm\theta$ in the target model, and the distributions and variables of the two models may be distinct as well. Actually, our models are of both
conditional distribution drift and covariate shift, and the difference between the two models may be quite significant. Thus, it is desired to develop a new transfer learning method for such heterogeneous models.

\subsection{Artificial accurate relationships between the models }

In order to construct the cr-TLL,
we first establish the relationships between the parameter vectors $\bm\beta$ and $(\bm\gamma,\bm\theta)$. By the expectation property of random variables, we have the following representations:
\begin{eqnarray}
E[Y^P X^P]= E\{X^P E[Y^P|X^P]\} \ \mbox{ and } \ E[Y^Q X^Q]= E\{X^Q E[Y^Q|X^Q,Z^Q]\}.
\end{eqnarray}
It reveals a simple and important phenomenon that the indices of linear correlations $E[Y^P X^P]$ and $E[Y^Q X^Q]$ can accurately evaluate the average effects of regressions $E[Y^P|X^P]$ and $E[Y^Q|X^Q,Z^Q]$ respectively in the two models. We then use the indices $E[Y^P X^P]$ and $E[Y^Q X^Q]$ to build the relationships between the parameters in the two models.
Let $X^{jP}$ and $X^{jQ}$ be the $j$-th components of $X^P$ and $X^Q$ respectively, and write
$$\Lambda =\mbox{diag}\left(\frac{E[Y^PX^{1P}]}{E[Y^QX^{1Q}]},\cdots,
\frac{E[Y^PX^{d_1P}]}{E[Y^QX^{d_1Q}]}\right).$$
Note that the indices $E[Y^PX^{jP}]$ and $E[Y^QX^{jQ}]$ or their ratios $\frac{E[Y^PX^{jP}]}{E[Y^QX^{jQ}]}$ can be easily estimated (see the Appendix). We then suppose that $\Lambda$ satisfies the following condition.
\begin{itemize} \item [\it C0.] {\it The matrix $\Lambda$ has been estimated by historical data or a part sample of $(X^P,Y^P)$ and $(X^Q,Y^Q)$.}\end{itemize}
Actually, this is the only special condition required in our methodology.
For describing the application background of this assumption, we introduce the following motivating examples of data condition from the perspectives of data streams and medical studies (for the motivating example on privacy policy see the Appendix):

{\it Example 1.} An interesting example comes from the area of data streams. It is known that the raw historical data of $(X^P,Y^P)$ and $(X^Q,Y^Q)$ may not be stored in the procedure of online updating because of huge amount of data in data streams, but some historical characteristics, such as the expectations or average values of $Y^P X^P$ and $Y^Q X^Q$, can be easily stored. Thus, we directly use the stored historical characteristics to construct $\Lambda$. It is worth pointing out that the stored characteristics of $(X^Q,Y^Q)$ cannot be directly used for inferring model $Q$ by commonly used methods, because model $Q$ contains an additional covariate $Z^Q$.

{\it Example 2.} The historical data volume of $(X^P,Y^P)$ and $(X^Q,Y^Q)$ can be large. As shown at the beginning of this section, in the procedure of data collection, the only difficulty is to collect the data of $Z^Q$. Then, the historical data of $Z^Q$ may be unavailable, or the data volume of $Z^Q$ is relatively small. In the example of U.S. carriers aforementioned in Introduction, the historical data of causes of flight delays (denoted by $Z^Q$) were not available before June 2003, but the historical data of the other relevant variables $(X^P,Y^P)$ and $(X^Q,Y^Q)$ were available before that time.
However, when the goal is to estimate the parameters (instead of $\Lambda$) in U.S. carrier model (denoted by $Q$) after June 2003, the data of $Z^Q$ are available. In this case, model $Q$ can be an updated version of $P$ by adding newly emerging variable $Z^Q$ after June 2003. As shown in Introduction, we call $(X^P,Y^P)$ and $(X^Q,Y^Q)$ permanent variables, and call $Z^Q$ emerging variable. Under such a situation, the historical data of  permanent variables can be used to estimate $\Lambda$. This example will be further discussed for actual data analysis in Section 5.

{\it Example 3.}
The above motivating examples mainly come from the area of data streams.  Actually, in regression settings, the emergence of new variables is very common due to some special causes, such as negligence in data collection under some cases, change of protocol, or advances in technology. In medical studies, for example, new medical devices can monitor measurements that have never been seen before, and by the new measurements, a new covariate vector, denoted by $Z^Q$, may emerge, and its observational values can be obtained. Under such a situation, the historical data or auxiliary
sample of $(X^P,Y^P)$ and $(X^Q,Y^Q)$  may be abundant, but the data size of $Z^Q$ is small, see, e.g.,  Hood et al. (2004). We then use the historical data or auxiliary
sample of $(X^P,Y^P)$ and $(X^Q,Y^Q)$ to estimate $\Lambda$.

For more similar examples see for example   Certo (2003) and Desyllas and Sako (2013).
With the above data condition, by likelihood equations (2.2) and (2.4), a preliminary but crucial relationship is given in the following lemma.

\noindent{\bf Lemma 2.1.} {\it If $E[Y^QX^{jQ}]\neq 0$ for $j=1,\cdots,d_1$, and $\Lambda$ satisfies the condition C0, then, the model parameters satisfy
\begin{eqnarray*}E[X^P \dot{G}_P(\bm\beta^T X^P)]=\Lambda  E[X^Q \dot{G}_Q(\bm\gamma^TX^Q+
\bm\theta^T Z^Q)].\end{eqnarray*}
\vspace{-3ex}}

The proof of the lemma is given in the Appendix.
The condition $E[Y^QX^{jQ}]\neq 0$ for $j=1,\cdots,d_1$, is reasonable. It is because if $E[Y^QX^{j^*Q}]=0$ for an index $j^*$, the corresponding component should be deleted from the likelihood equation (2.4) before any inference. For better understanding the lemma, we introduce the following explanation.

\noindent{\bf Remark 2.1.} {\it (i) It follows from (2.6) and the expectation property of exponential family distribution that the formula in Lemma 2.1 can be recast as
\begin{eqnarray}
E\{X^P E[Y^P|X^P]\}=\Lambda  E\{X^Q E[Y^Q|X^Q,Z^Q]\}.
\end{eqnarray}
Actually, the formula (2.7) precisely captures the relationship between the weighted averages of the regressions respectively in models $P$ and $Q$, in other words, the two regressions are precisely combined artificially by the diagonal matrix $\Lambda$. This is a bran-new measure for the relationship between the two models. Here $\Lambda$ could be thought of as a correlation-ratio of the linear correlations between the responses and covariates respectively in the two models. We then call it the correlation-ratio matrix. The matrix measures the ``linear correlation ratio" between the models, and based on the structural information of exponential family distribution, the matrix precisely establishes the relationship between the parameters in the two models.

(ii) It is worth pointing out that as shown above, the lemma presents the artificial relationship between the regressions, instead of the similarity between the source and target. Without the similarity measure, we can still transfer the information across the tasks; the issue will be further discussed in Remark 3.1 and Remark 3.2.

(iii) The formula in the lemma is derived from the likelihood equations (2.2) and (2.4). It will be seen in the following discussion that this way can result in a favorable estimator of $\bm\gamma$, instead of $\bm\theta$, because the likelihood equations (2.2) and (2.4) are designed only for $\bm\beta$ and $\bm\gamma$ respectively, but not for $\bm\theta$.  Similarly, we can use the equation (2.2) together with (2.5), instead of (2.4), to construct the linear correlation-ratio. By this treatment, the method can improve the estimator of $\bm\theta$, rather than $\bm\gamma$. In the following, we mainly focus on the relationship between $\bm\beta$ and $\bm\gamma$ derived from the equations (2.2) and (2.4).

}

Note that we need to express $\bm\beta$ as a function of $\bm\gamma$ and $\bm\theta$, instead of its reversion,  because we want to express the likelihood of model $P$ as a function of $\bm\gamma$ and $\bm\theta$; for the details see the next subsection. Lemma 2.1 shows that the parameter vector $\bm\beta$ in model $P$ is a function of the parameter vectors $\bm\gamma$ and $\bm\theta$ in model $Q$. The existence theorem of implicit functions
can ensure that such a function is existent and unique under some
regularity conditions.

Particularly, for linear regressions, we have the following simple representations.

\noindent{\bf Theorem 2.2.} {\it If model $P$ is a linear regression, $E[Y^QX^{jQ}]\neq  0$ for $j=1,\cdots,d_1$, $(E[X^P(X^P)^T])^{-1}$ exists, and $\Lambda$ satisfies the condition C0, then, the model parameters satisfy
\begin{eqnarray*}\bm\beta=(E[X^P(X^P)^T])^{-1}\Lambda E[X^Q \dot{G}_Q(\bm\gamma^TX^Q+
\bm\theta^T Z^Q)].\end{eqnarray*}
In addition to the above conditions, if model $Q$ is also a linear regression, then, the model parameters have the following linear relationship:
\begin{eqnarray*}\bm\beta=(E[X^P(X^P)^T])^{-1}\Lambda \left(E[X^Q(X^Q)^T]\bm\gamma
+E[X^Q(Z^Q)^T]\bm\theta\right).\end{eqnarray*} \vspace{-4ex}
}

We make a remark on the relationships proposed in the theorem.

\noindent{\bf Remark 2.2.} {\it The theorem shows that if only $P$ is linear, the parameter vector $\bm\beta$ can be expressed as a nonlinear function of $(\bm\gamma,\bm\theta)$, and for the case where the two models are linear, the relationship between the parameters in the two models is linear as well via correlation-ratio matrix $\Lambda$. }

In the following, we consider the case of non-linear GLMs. Under the situation, also we can use the implicit function given in Lemma 2.1, and its iterative algorithms to determine the relationship between $\bm\beta$  and $(\bm\gamma,\bm\theta)$ (see the Appendix). For computational efficiency, however, here we still explore an explicit expression for the relationship. Denote by $\ddot{G}_P(\cdot)$ and $\ddot{G}_Q(\cdot)$ the second-order derivatives of $G_P(\cdot)$ and $G_Q(\cdot)$, respectively. Suppose that $\ddot G_P(u)$ and $\ddot G_Q(u)$ are absolutely continuous functions, and $E[|\ddot G_P(\bm\beta^TX^P)|]$ and $E[|\ddot G_Q(\bm\gamma^TX^Q+\bm\theta^TZ^Q)|]$ exist for $\bm\beta\in\cal{B}$, $\bm\gamma\in\Gamma$ and $\bm\theta\in\Theta$, where $\cal B$, $\Gamma$ and $\Theta$ are parameter spaces of $\bm\beta$, $\bm\gamma$ and $\bm\theta$, respectively. By Lemma 2.1 and the First-Order Stein's Identity (Stein, 1986; Yang et al., 2017), we have the following theorem.

\noindent{\bf Theorem 2.3.} {\it If both $P$ and $Q$ are non-linear GLMs, $E[Y^QX^{jQ}]\neq  0$ for $j=1,\cdots,d_1$, $E\{Var[Y^P|X^P]\}\neq 0$, $X^P\sim N(0,I_{d_1})$, and $\Lambda$ satisfies the condition C0, then, the model parameters satisfy
\begin{eqnarray*}\bm\beta=\frac{1}{E\{Var[Y^P|X^P]\}}
\Lambda\, E[X^Q\dot G_Q(\bm\gamma^TX^Q+\bm\theta^TZ^Q)].\end{eqnarray*} Addition to the above conditions, if $X^Q\sim N(0,I_{d_1})$, then
\begin{eqnarray*}\bm\beta=\frac{E[\ddot{G}_Q(\bm\gamma^T X^Q+\bm\theta^T Z^Q)]}{E\{Var[Y^P|X^P]\}}
\Lambda\, \bm\gamma.\end{eqnarray*}\vspace{-3ex}}

For the theorem, we have the following explanations.

\noindent{\bf Remark 2.3.} {\it The above are non-linear relationships between the parameters. In the theorem, we need to estimate $E\{Var[Y^P|X^P]\}$ to quantify the relationships between the parameters. The estimator can be accurate because the sample size of model $P$ can be large. However, the theorem needs the condition that the involved covariates obey standard normal distributions. Without the normality, it will be seen that the relationship is relatively complex (see the Appendix).}

For general complex cases, addition to the iterative algorithms in the Appendix, if some historical characteristics of $Z^Q$ are also available, an explicit expression of the function relationship can be attained, for the details see  Theorem A.2 in the Appendix.

\subsection{Transfer learning estimators via cr-TLL }

Based on the relationships proposed above, we can establish the cr-TLLs for target parameters.
Suppose that we observe $n_P$ independent and identically distributed (i.i.d.) samples $(X^P_1,Y^P_1),\cdots,(X^P_{n_P},Y^P_{n_P})$ drawn from the source model $P$, and $n_Q$ i.i.d. samples $((X^Q_1,Z^Q_1),Y^Q_1),\cdots,((X^Q_{n_Q},Z^Q_{n_Q}),Y^Q_{n_Q})$ drawn from the target model $Q$. The data points from the models $P$ and $Q$ are
also mutually independent, and the sample sizes satisfy $n_P\geq n_Q$.

According to Theorem 2.2 and Theorem 2.3 given above, and Theorem A.2 and the iterative algorithms given in the Appendix, these accurate and approximate representations can be unified denoted by
\begin{eqnarray}
\bm\beta=\bm s(\bm\gamma,\bm\theta).
\end{eqnarray}
By this notation, under GLMs, the accurate log-likelihood from the two models can be together expressed as
\begin{eqnarray}
\nonumber l_{(P,Q)}(\bm\gamma,\bm\theta)&=&w_P\sum_{i=1}^{n_P}\left(Y^P_is^T(\bm\gamma,\bm\theta) X^P_i -G_P(s^T(\bm\gamma,\bm\theta) X^P_i)\right)\\&&+w_Q\sum_{i=1}^{n_Q}\left(Y_i^Q(\bm\gamma^TX^Q_i+
\bm\theta^T Z^Q_i) -G_Q(\bm\gamma^TX^Q_i+
\bm\theta^T Z^Q_i)\right),
\end{eqnarray}
 where $w_P$ and $w_Q$ are weights satisfying $w_P>0$, $w_Q>0$ and $w_P+w_Q=1$. In GLMs, the weights $w_P$ and $w_Q$ should be inversely proportional to the dispersion parameters of model $P$ and model $Q$, respectively. Thus, if the dispersion parameters are unknown, the weights $w_P$ and $w_Q$ can be estimated by for example $\widehat w_P=\frac{w_1}{w_1+w_2}$ and $\widehat w_Q=\frac{w_2}{w_1+w_2}$ respectively, where
\begin{eqnarray}
 && w_1=\frac{1}{\frac{1}{n_P-1}\sum_{i=1}^{n_P}(Y_i^P-
g_P^{-1}({\bm\beta^*}^{T}X_i^P))^2},\\
\label{eqn_weight_2}
&& w_2=\frac{1}{\frac{1}{n_Q-1}\sum_{i=1}^{n_Q}(Y_i^Q-g_Q^{-1}
({\bm\gamma^*}^{T}X_i^Q+
{\bm\theta^*}^{T}Z_i^Q))^2},
\end{eqnarray}
 and $\bm\beta^*$, $\bm\gamma^*$ and $\bm\theta^*$ are the consistent estimators of $\bm\beta$, $\bm\gamma$ and $\bm\theta$ by the data from $P$ and $Q$, respectively. Generally, the choices of weights are not important. We can choose the weight estimators $\widehat w_P$ and $\widehat w_Q$ satisfying $\widehat w_P=w_p+o_P(1)$ and $\widehat w_Q=w_Q+o_P(1)$ for some deterministic quantities $w_P$ and $w_Q$.

The above likelihood only contains the parameters of interest $(\bm\gamma,\bm\theta)$, and all the data from both models $P$ and $Q$ via correlation-ratio combination.  We then call it correlation-ratio
transfer learning likelihood (denoted by cr-TLL for short).
In the cr-TLL (2.9), however, the representation $\bm s(\bm\gamma,\bm\theta)$ given in (2.8) depends on the estimators of $\Lambda$, $E[X^P(X^P)^T]$, $E[X^Q(X^Q)^T]$ and so on. We first estimate $\Lambda$ and $E[X^P(X^P)^T]$ via historical data or historical characteristics, and estimate $E[X^Q(X^Q)^T]$ by current sample. Then, we attain the consistent estimator $\widehat {\bm s}(\bm\gamma,\bm\theta)$ of $\bm s(\bm\gamma,\bm\theta)$ satisfying
\begin{eqnarray}\label{(representation-1_1)}
\widehat {\bm s}(\bm\gamma,\bm\theta)=\bm s(\bm\gamma,\bm\theta)+O_p\left(1/\sqrt{n_m}\right) \ \mbox{ uniformly for } \bm\gamma\in \Gamma \mbox{ and }\bm\theta\in\Theta,\end{eqnarray} where $n_m=\min\{\widetilde n,n_Q\}$ with $\widetilde n$ being the sample size of historical data. We need the condition $\widetilde n\rightarrow\infty$ as $n_Q\rightarrow\infty$ for theoretical analysis.
Consequently, the estimated cr-TLL can be expressed as
\begin{eqnarray}
\nonumber\hspace{-0.7cm}\widehat l_{(P,Q)}(\bm\gamma,\bm\theta)&=&\widehat w_P\sum_{i=1}^{n_P}\left(Y^P_i\widehat {\bm s}^T(\bm\gamma,\bm\theta) X^P_i -G_P(\widehat {\bm s}^T(\bm\gamma,\bm\theta) X^P_i)\right)\\&& +\widehat w_Q\sum_{i=1}^{n_Q}\left(Y_i^Q(\bm\gamma^TX^Q_i+
\bm\theta^T Z^Q_i) -G_Q(\bm\gamma^TX^Q_i+
\bm\theta^T Z^Q_i)\right).
\end{eqnarray}
Finally, the transfer learning likelihood estimator is obtained as
\begin{eqnarray}
\left(\begin{array}{ll}\widehat{\bm\gamma}\\\widehat{\bm\theta}\end{array}
\right)
= \arg\max_{\bm\gamma\in \Gamma,\bm\theta\in \Theta} \widehat l_{(P,Q)}(\bm\gamma,\bm\theta).\end{eqnarray}

To illustrate the above transfer learning likelihood estimator, consider the simple scenario where $P$ and $Q$ are linear regression models with $d_1=d_2=1$, and all the covariates are centralized and standardized such that $E[X^P]=E[X^Q]=E[Z^Q]=0$ and $E[(X^P)^2]=E[(X^Q)^2]=E[(Z^Q)^2]=1$. It follows from Theorem 2.2 that the estimator of ${\bm s}(\bm\gamma,\bm\theta)$ can be expressed as
\begin{eqnarray}\label{(a-b-estimation)}\widehat{{\bm s}}(\bm\gamma,\bm\theta)=\widehat a\bm\gamma+\widehat b\bm\theta\triangleq  \widehat{ \left(\frac{E[Y^PX^P]}{E[Y^QX^Q]}\right)}\bm\gamma+
\widehat{\left(\frac{E[Y^PX^P]}{ E[Y^QX^Q]}\right)}\widehat E[X^QZ^Q]\bm\theta,\end{eqnarray} where the notation $ \left(\widehat {\,\frac{ \,\, \,\,\cdot \,\,\,\,}{\,\, \,\,\cdot \, \,\,\, } \,}\right)$ stands for the bias-correction estimator of the ratio of the parameters by historical data or historical characteristics, and $\widehat E[X^QZ^Q]$ is an empirical estimator of $E[X^QZ^Q]$; the estimation details are given in the Appendix. From (2.13), the corresponding likelihood equation is
\begin{eqnarray*}&&\widehat w_P\sum_{i=1}^{n_P}\left(Y^P_iX^P_i
\left(\begin{array}{ll}\widehat a\\ \widehat b\end{array}\right) -(X_i^P)^2\left(\begin{array}{ll}\widehat a^2&\widehat a\,\widehat b\\ \widehat a\,\widehat b&\widehat b^2\end{array}\right) \left(\begin{array}{ll}\bm\gamma\\\bm\theta\end{array}\right)\right)\\ && +\widehat w_Q\sum_{i=1}^{n_Q}\left(Y_i^Q\left(\begin{array}{ll} X^Q_i\\ Z^Q_i\end{array}\right) -\left(\begin{array}{ll} (X^Q_i)^2&X_i^QZ_i^Q\\ X^Q_iZ^Q_i&(Z_i^Q)^2\end{array}\right) \left(\begin{array}{ll}\bm\gamma\\\bm\theta\end{array}\right)\right)=0.
\end{eqnarray*} We then get the transfer learning estimators as
\begin{eqnarray}\nonumber
\left(\begin{array}{ll}\widehat{\bm\gamma}
\\\widehat{\bm\theta}\end{array}
\right)&=&\left(\widehat w_P\sum_{i=1}^{n_P}(X_i^P)^2\left(\begin{array}{ll}\widehat a^2&\widehat a\,\widehat b\\ \widehat a\,\widehat b&\widehat b^2\end{array}\right)+\widehat w_Q\sum_{i=1}^{n_Q}\left(\begin{array}{ll} (X^Q_i)^2&X_i^QZ_i^Q\\ X^Q_iZ^Q_i&(Z_i^Q)^2\end{array}\right)\right)^{-1}\\&&\times\left(
\widehat w_P\sum_{i=1}^{n_P}Y^P_iX^P_i
\left(\begin{array}{ll}\widehat a\\ \widehat b\end{array}\right) +\widehat w_Q\sum_{i=1}^{n_Q}Y_i^Q\left(\begin{array}{ll} X^Q_i\\ Z^Q_i\end{array}\right)\right).\end{eqnarray} The estimators use all the data from two models $P$ and $Q$, and the expression form (2.13) is similar to the original least squared estimator. The favorable property of the estimators will be given in the next section.

\subsection{Extension to the case of multiple source models}

To extend the strategy proposed to the case of multiple source models, we suppose that there are $m$ source models, denoted by $P_j$ for $j=1,\cdots,m$, that are of framework of GLMs, and are distinct from the target model $Q$. Under the source model $P_j$, the related regression coefficient vector is written by $\bm\beta_j$  and the corresponding relationship between $\bm\beta_j$ and $(\bm\gamma,\bm\theta)$ is then written as
$\bm\beta_j=\bm s_j(\bm\gamma,\bm\theta)$.

With the above notations, the estimated cr-TLL can be expressed as
\begin{eqnarray}\label{(multi-likelihood)}\nonumber\hspace{-0.7cm}&&\widehat l_{((P_1,\cdots,P_m),Q)}(\bm\gamma,\bm\theta)\\&&\nonumber=\sum_{j=1}^m\widehat w_{P_j}\sum_{i=1}^{n_{P_j}}\left(Y^{P_j}_i\widehat {\bm s}_j^T(\bm\gamma,\bm\theta) X^{P_j}_i -G_{P_j}(\widehat {\bm s}_j^T(\bm\gamma,\bm\theta) X^{P_j}_i)\right)\\&& \ \ \ \ +\widehat w_Q\sum_{i=1}^{n_Q}\left(Y_i^Q(\bm\gamma^TX^Q_i+
\bm\theta^T Z^Q_i) -G_Q(\bm\gamma^TX^Q_i+
\bm\theta^T Z^Q_i)\right),\end{eqnarray} where $(X^{P_j}_1,Y^{P_j}_1),\cdots,(X^{P_j}_{n_{P_j}},Y^{P_j}_{n_{P_j}})$ are independent and identically distributed samples drawn from $P_j$ respectively for $j=1,\cdots,m$, and $\widehat w_{P_1},\cdots,\widehat w_{P_m}$, are the estimated weights satisfying the conditions similar to those given in (2.10) and (2.11). The above is the extended version of cr-TLL (2.13).
Finally, the transfer learning likelihood estimators are obtained as
\begin{eqnarray}\label{(multi-estimator)}
\left(\begin{array}{ll}\widehat{\bm\gamma}\\\widehat{\bm\theta}\end{array}
\right)
= \arg\max_{\bm\gamma\in \Gamma,\bm\theta\in \Theta} \widehat l_{((P_1,\cdots,P_m),Q)}(\bm\gamma,\bm\theta).\end{eqnarray}

Consider a simple scenario where $P_j$ and $Q$ are linear regression models with $d_1=d_2=1$, and all the cavariates are centralized and standardized such that $E[X^{P_j}]=E[X^Q]=E[Z^Q]=0$ and $E[(X^{P_j})^2]=E[(X^Q)^2]=E[(Z^Q)^2]=1$. Similar to (2.13), the transfer learning estimator can be written as
\begin{eqnarray}\nonumber\label{(multi-LS)}
\left(\begin{array}{ll}\widehat{\bm\gamma}
\\\widehat{\bm\theta}\end{array}
\right)&=&\left(\sum_{j=1}^{m}\widehat w_{P_j}\sum_{i=1}^{n_{P_j}}(X_i^{P_j})^2\left(\begin{array}{cc}\widehat a^2_j&\widehat a_j \widehat b_j\\ \widehat a_j \widehat b_j&\widehat b^2_j\end{array}\right)+\widehat w_Q\sum_{i=1}^{n_Q}\left(\begin{array}{ll} (X^Q_i)^2&X_i^QZ_i^Q\\ X^Q_iZ^Q_i&(Z_i^Q)^2\end{array}\right)\right)^{-1}\\&&\times\left(
\sum_{j=1}^{m}\widehat w_{P_j}\sum_{i=1}^{n_{P_j}}Y^{P_j}_iX^{P_j}_i
\left(\begin{array}{ll}\widehat a_j\\ \widehat b_j\end{array}\right) +\widehat w_Q\sum_{i=1}^{n_Q}Y_i^Q\left(\begin{array}{ll} X^Q_i\\ Z^Q_i\end{array}\right)\right),\end{eqnarray} where
$$\widehat a_j=\widehat{\left(\frac{E[Y^{P_j}X^{P_j}]}{E[Y^QX^Q]}\right)} \ \mbox{ and } \ \widehat b_j=
\widehat{\left(\frac{E[Y^{P_j}X^{P_j}]}{ E[Y^QX^Q]}\right)}\widehat E[X^QZ^Q].$$

\setcounter{equation}{0}
\section{Asymptotic properties and key conditions}

For the sake of simplicity of theoretical statements, we in this section suppose that the limit $\lim\limits_{n\rightarrow\infty}\frac{n_Q}{n_P}$ exists, denoted by   $\tau=\lim\limits_{n\rightarrow\infty}\frac{n_Q}{n_P}$.

We begin with discussing the transfer learning estimators (2.13) under simple linear regressions, in which the covariates are real-valued, and are centralized and standardized.  We compare the estimator (2.13) with the Least Squares (LS) estimators $\widehat{\bm\gamma}^Q$ and $\widehat{\bm\theta}^Q$ obtained only by the data in the target model $Q$. Let $n=n_P+n_Q$,
$\rho_{X^QZ^Q}$ be the correlation coefficient between $X^Q$ and $Z^Q$, and
\begin{eqnarray}\label{(notation)}(a,b)=\left(\frac{E[Y^PX^P]}{E[Y^QX^Q]},
\frac{E[Y^PX^P]}{E[Y^QX^Q]}E[X^QZ^Q]\right).\end{eqnarray}

It follows from the proof of Theorem 3.1 that the transfer learning estimators in (2.13) are approximately unbiased, i.e.,
$$E\left[\left(\begin{array}{ll}\widehat{\bm\gamma}
\\ \widehat{\bm\theta}\end{array}
\right)\right]\approx\left(\begin{array}{ll}{\bm\gamma}
\\ {\bm\theta}\end{array}
\right),$$
and, asymptotically, the covariance satisfies
\begin{eqnarray*} Cov\left[\left(\begin{array}{ll}\widehat{\bm\gamma}
\\\widehat{\bm\theta}\end{array}
\right)\right]&\approx &\left(n_Pw_P\left(\begin{array}{cc}a^2& ab\\ ab& b^2\end{array}\right)+n_Qw_Q\left(\begin{array}{cc} 1&\rho_{X^QZ^Q}\\ \rho_{X^QZ^Q}&1\end{array}\right)\right)^{-1}.\end{eqnarray*} Furthermore, for the convenience of representation, we consider the simple case where the two models $P$ and $Q$ have the same error variance $\sigma_P^2=\sigma_Q^2=1$. We then set weights by $\widehat w_P=\widehat w_Q=1/2$. The following theorem presents the key conditions and states the main theoretical conclusions for linear models.

\noindent{\bf Theorem 3.1.} {\it Under the conditions of Theorem 2.2, suppose that linear models models $P$ and $Q$ satisfy the above conditions, $|\rho_{X^QZ^Q}|<1/\sqrt 2$ and the following condition  holds:
\begin{itemize} \item [C1.]  $a^2>\tau\rho^2_{X^QZ^Q}/(1-2\rho^2_{X^QZ^Q}).$
\end{itemize}
(1) Then, the transfer learning estimator $\widehat{\bm\gamma}$ in (2.13) has the following asymptotic normality:
$$\sqrt {v_{\bm\gamma}} (\widehat{\bm\gamma}-{\bm\gamma}^0)\rightarrow_d N(0,1)  \mbox{ as } n_P, n_Q \mbox{ and } \widetilde n\rightarrow\infty,$$ where $\widetilde n$ is the size of historical sample,
$v_{\bm\gamma}=n_P \varphi(a,\tau,\rho)+n_Q>0$ with $$\varphi(a,\tau,\rho)=a^2\left(1-\frac{\rho^2_{X^QZ^Q}\left
(1+\frac{\tau}{a^2}\right)^2}
{\rho^2_{X^QZ^Q}+\frac{\tau}{a^2}}\right).$$

\noindent
(2) Particularly, if
\begin{itemize}
 \item [C2.] $\rho_{X^QZ^Q}\neq 0$ and $\tau\neq 0$,
\end{itemize}
then, $\varphi(a,\tau,\rho)>0$. Consequently,
the convergence rate of $\widehat{\bm\gamma}$ in probability can be expressed as
\begin{eqnarray*}\sqrt {v_{\bm\gamma}}=\sqrt{n_P \varphi(a,\tau,\rho)+n_Q}
=O\left(\sqrt n\right).
\end{eqnarray*}

\noindent (3) Contrarily, if
\begin{itemize}
 \item [C3.] $\rho_{X^QZ^Q}= 0$ and $\tau\geq0$,
\end{itemize} then, the convergence rate of $\widehat{\bm\gamma}$ in probability can be expressed as
\begin{eqnarray*}\sqrt {v_{\bm\gamma}}=\sqrt{a^2n_P +n_Q}
=O\left(\sqrt n\right).
\end{eqnarray*}

}

From the theorem, we have the following observations.

\noindent{\bf Remark 3.1.} {\it (i) Note that $a=\frac{E[Y^PX^P]}{E[Y^QX^Q]}$ by the definition. Thus, the key condition C1 means that the linear correlation between $X^P$ and $Y^P$ is not much weak compared to the linear correlation between $X^Q$ and $Y^Q$.
With the condition, the use of the information of $P$ can achieve the global convergence rate of order $\sqrt{n}$ in probability.
However, the LS estimator $\widehat{\bm\gamma}^Q$ in model $Q$ only has the local convergence rate of order $\sqrt{n^Q}$.

(ii) It is can be verified that the function $\varphi(a,\tau,\rho)$ satisfies $\varphi(-a,\tau,\rho)=\varphi(a,\tau,\rho)$, and its derivative with respect to $a$ has the form $\dot\varphi(a,\tau,\rho)
=2a\left(\frac{\tau\left(1-\rho^2_{X^QZ^Q}\right)}
{a^2\rho^2_{X^QZ^Q}+\tau}\right)^2$ and satisfies $\dot\varphi(a,\tau,\rho)\geq 0$ if $a\geq 0$, $\dot\varphi(a,\tau,\rho)< 0$ if $a< 0$, and $\lim\limits_{|a|\rightarrow +\infty}\dot\varphi(a,\tau,\rho)=0$. These show that relatively large values of $|a|$ can enhance the consistency of the estimator, implying that the strong linear correlation between $X^P$ and $Y^P$ helps to improve the estimator.

(iii) Note that the key condition C1 does not imply the similarity between $P$ and $Q$. Thus, it is interesting that the phenomenon of a variational Stein's paradox (see, e.g.,  Stein et al., 1956; Stigler, 1990) can be seen from the condition and the theoretical conclusions. In other words, even the source model is unrelated to the target model in some sense, the information of the source model can help to improve the estimation accuracy for the parameters in the target model. Although the historical data or characteristics are used to combine the two models, the historical data in the two models are also independent of each other. Thus, the model conditions are basically in accord with those in
Stein's paradox (for the details see the Appendix).

(iv) The key condition C2 implies that if models $P$ and $Q$ are not similar, it is impossible to immoderately enhance the convergence rate via increasing the sample size $n_P$, because $n_P$ and $n_Q$ are of infinity of the same order. Actually, by the condition, the second result of the theorem implies that
the convergence rate of $\widehat{\bm\gamma}$ in probability is
\begin{eqnarray*}\sqrt {v_{\bm\gamma}}=\sqrt{n_P \varphi(a,\tau,\rho)+n_Q}
=\sqrt{cn_Q}
\end{eqnarray*} for some constant $c\approx \sqrt{\tau^{-1} \varphi(a,\tau,\rho)+1}>1$. This means that although the transfer learning estimator can achieve $\sqrt n$-consistency, the achieved convergence rate is only of the multiply of $\sqrt {n_Q}$.

(v) Contrarily, in the key condition C3, $\rho_{X^QZ^Q}=0$ implies that the models are somewhat similar. Then, without the constraint on $\tau$, the convergence rate of $\widehat{\bm\gamma}$ is \begin{eqnarray*}\sqrt {v_{\bm\gamma}}=\sqrt{a^2n_P +n_Q}.
\end{eqnarray*} Particularly, when models $P\approx Q$, then $a\approx 1$, the convergence rate is thus $$\sqrt {v_{\bm\gamma}}\approx\sqrt{n_P +n_Q}=\sqrt{n}. $$
Therefore, in the case of $\rho_{X^QZ^Q}=0$, the convergence rate can be enhanced by increasing the sample size of model $P$.

}

These observations are not incompatible with the no-free-lunch theory in   Hanneke and Kpotufe (2020). For comparing the estimators $\widehat{\bm\gamma}$ and $\widehat{\bm\theta}$, we have the following consequence.

\noindent{\bf Proposition 3.2.} {\it Under the conditions of Theorem 3.1, the transfer learning estimator $\widehat{\bm\theta}$ in (2.13) satisfies
$$\sqrt {v_{\bm\theta}} (\widehat{\bm\theta}-{\bm\theta}^0)\rightarrow_d N(0,1) \mbox{ as } n_P, n_Q \mbox{ and } \widetilde n\rightarrow\infty,$$ where
$v_{\bm\theta}=n_Q(1-\rho_{X^QZ^Q}),$ implying that the convergence rate in probability is
$\sqrt {v_{\bm\theta}}=\sqrt{n_Q(1-\rho_{X^QZ^Q})},$ which is slower than  $\sqrt {v_{\bm\gamma}}$. }

The proposition shows that the convergence rate of the estimator of parameter vector $\bm\theta$ cannot be enhanced by the linear correlation-ratio derived from the equations (2.2) and (2.4). Actually, in contrast with Theorem 3.1, we have that  if the linear correlation-ratio is derived from the equations (2.2) and (2.5), the convergence rate of the estimator of parameter vector $\bm\theta$ can be enhanced by the transfer learning method.

In the following, we consider the theoretical property for general case. Suppose that $E[X^P]=0$, $E[X^Q]=0$, $E[Z^Q]=0$, and their components $X^{jP}$, $X^{jQ}$ and $Z^{jQ}$ standardized satisfying $Var[X^{jP}]=1$, $Var[X^{jQ}]=1$ and $Var[Z^{jQ}]=1$, without loss of generality.
We first introduce the following notations: Let $\bm\alpha=(\bm\gamma^T,\bm\theta^T)^T$, $\widehat{\bm\alpha}=(\widehat{\bm\gamma}^T,\widehat{\bm\theta}^T)^T$, $\bm\alpha^0$ be the true value of $\bm\alpha$, and $\dot{\bm s}(\bm\alpha)$ be the $d_1\times (d_1+d_2)$-dimensional derivative matrix of $\bm s(\bm\alpha)$.
Write
\begin{eqnarray*}
&&Var[Y_i^P]=\sigma_{P, i}^2,~~~~ Var[Y_i^Q]=\sigma_{Q, i}^2,\\
&&F^P_{n_P}(\bm\alpha)= \dot{\bm s}^T(\bm\alpha)\sum_{i=1}^{n_P}\sigma_{P, i}^2X_i^{P}(X_i^{P})^T\dot{\bm s}(\bm\alpha),\\
&&F^Q_{n_Q}(\bm\alpha)=\sum_{i=1}^{n_Q}\sigma_{Q, i}^2((X_i^Q)^T,(Z_i^Q)^T)^T((X_i^Q)^T,(Z_i^Q)^T),\\
&&F_{n}(\bm\alpha)=w_PF^P_{n_P}(\bm\alpha)+w_QF^Q_{n_Q}(\bm\alpha).
\end{eqnarray*}

For establishing asymptotic theory, the required regularity conditions C6-C10 are presented in the Appendix. We are now in the position to give general theoretical conclusions.

\noindent {\bf Theorem 3.3.} {\it Under the conditions of Theorem 2.3 and the regularity conditions C6-C10 given in the Appendix,
then, the general transfer learning estimator in (2.14) is asymptotically normal as
$$F_n^{1/2}(\bm\alpha^0)(\widehat{\bm\alpha}-\bm\alpha^0)\rightarrow_d N(0,I)  \mbox{ as } n_P, n_Q \mbox{ and } \widetilde n\rightarrow\infty.\vspace{-3ex}$$}

For better understanding and explaining the theorem, we consider the case of $\widehat w_P=\widehat w_Q=1/2$, and suppose the following limits exist: \begin{eqnarray*}&&\frac{1}{n_P}\sum_{i=1}^{n_P}\sigma_{P, i}^2X_i^{P}
(X_i^{P})^T\rightarrow_p \Sigma^P \ \mbox{ as }n_P\rightarrow\infty, \\&& \frac{1}{n_Q}\sum_{i=1}^{n_Q}\sigma_{Q, i}^2((X_i^Q)^T,(Z_i^Q)^T)^T
((X_i^Q)^T,(Z_i^Q)^T)\rightarrow_p \Sigma^Q\ \mbox{ as }n_Q\rightarrow\infty,\end{eqnarray*} where $\Sigma^P$ and $\Sigma^Q$ are $d_1\times d_1$ and $(d_1+d_2)\times (d_1+d_2)$-dimensional positive definite matrices, respectively. By the theorem and the same argument as used in linear regression, we will verify that every  component of the transfer learning estimator $\widehat{\bm\gamma}$ can achieve $\sqrt{n}$-consistency under some regularity conditions.

As an example, the convergence rate of the estimator $\widehat{\bm\gamma}_1$ of the first component of $\widehat{\bm\gamma}$ is presented in the following corollary. To this end, we first introduce the following notations. Denote $\dot{\bm s}(\bm\alpha^0)=(\dot{\bm s}_1,\cdots,\dot{\bm s}_{d_1+d_2})$, $\dot{\bm s}_{(-1)}(\bm\alpha^0)=(\dot{\bm s}_2,\cdots,\dot{\bm s}_{d_1+d_2})$, $\Sigma^Q=(\sigma^Q_{ij})_{i,j=1}^{d_1+d_2}$, and $ \bm\sigma^Q_{12}=(\sigma^Q_{12},\cdots,\sigma^Q_{1(d_1+d_2)})^T$.
Rewrite the $(d_1+d_2)\times (d_1+d_2)$-dimensional matrices $(\dot{\bm s}_i^T\Sigma^P\dot{\bm s}_j)_{i,j=1}^{d_1+d_2}$ and $\Sigma^Q$ respectively as $$(\dot{\bm s}_i^T\Sigma^P\dot{\bm s}_j)_{i,j=1}^{d_1+d_2}=\left(\begin{array}{cc}\dot{\bm s}_1^T\Sigma^P\dot{\bm s}_1&\dot{\bm s}_1^T\Sigma^P\dot{\bm s}_{(-1)}\\\dot{\bm s}_{(-1)}^T\Sigma^P\dot{\bm s}_1& \dot{\bm s}_{(-1)}^T\Sigma^P\dot{\bm s}_{(-1)}\end{array}\right)\ \mbox{ and } \
\Sigma^Q=\left(\begin{array}{cc}
\sigma^Q_{11}&\bm (\bm\sigma^Q_{12})^T\\\bm \sigma^Q_{12}&\Sigma^Q_{22}\end{array}\right).$$
Furthermore, let
$r=(\dot{\bm s}_1^T\Sigma^P\dot{\bm s}_{(-1)}+\tau\bm (\bm\sigma^Q_{12})^T)( A+\tau \Sigma^Q_{22})^{-1}(\dot{\bm s}_{(-1)}^T\Sigma^P\bm \dot{\bm s}_1+\tau\bm \sigma^Q_{12})$, $
D=A^{+}-\tau A^{+}((\Sigma^Q_{22})^{-1}+\tau A^{+})^{-1} A^{+}$, where $A=\dot{\bm s}_{(-1)}^T\Sigma^P\dot{\bm s}_{(-1)}$ and the notation $A ^{+}$ stands for the Moore-Penrose generalized inversion of the matrix $A$.
We then have the following consequence.

\noindent {\bf Corollary 3.4.} {\it Under the conditions of Theorem 2.3 and the regularity conditions C6-C10 given in the Appendix,
then, the first component estimator $\widehat \gamma_1$ has the following asymptotic normality:
$$\sqrt {v_{\bm\gamma_1}} (\widehat{\bm\gamma}_1-{\bm\gamma}_1^0)\rightarrow_d N(0,1) \mbox{ as } n_P, n_Q \mbox{ and } \widetilde n\rightarrow\infty,$$ where
$v_{\bm\gamma_1}=n_P\dot{\bm s}^T_1\Sigma^P\dot{\bm s}_1+n_Q\sigma^Q_{11}-n_Pr$. In addition to the conditions above, suppose that the following conditions hold:
\begin{itemize}\item [C4.] $\dot{\bm s}_1\neq \bm 0$,
\item [C5.] $\tau\neq 0$ or $\tau\geq0$ and $\dot{\bm s}^T_1 A^{+}\dot{\bm s}_1>0$, \end{itemize}
and there exists a constant $\epsilon>0$ such that
\begin{equation}
\begin{split}&\bm (\bm\sigma^Q_{12})^T( A+\tau \Sigma^Q_{22})^{-1}\bm \sigma^Q_{12}<\epsilon\,\dot{\bm s}^T_1( A+\tau \Sigma^Q_{22})^{-1}\dot{\bm s}_1,\\&(\tau^2\epsilon+2\tau\sqrt \epsilon)\,\bm \dot{\bm s}_1^T\Sigma^P\dot{\bm s}_{(-1)} A^{+}\dot{\bm s}_{(-1)}^T\Sigma^P\bm \dot{\bm s}_1\leq\bm \dot{\bm s}_1^T\Sigma^P\dot{\bm s}_{(-1)}D\dot{\bm s}_{(-1)}^T\Sigma^P\bm \dot{\bm s}_1.\end{split}\end{equation}
Then, the asymptotic variance of the first component estimator $\widehat{\bm\gamma}_1$ satisfies
$$v_{\bm\gamma_1}=O(1/n).\vspace{-3ex}$$
}

Note that the matrix
$( A+\tau \Sigma^Q_{22})^{-1}$ is positive definite. Thus, when the correlation coefficient vector $\bm \sigma^Q_{12}$ is close to a zero vector and/or the ``signal" in the vector ${\bm s}_1$ is strong, the condition (3.2) can be easily satisfied.
For further explaining Theorem 3.3 and Corollary 3.4, we give the following remark.

\noindent{\bf Remark 3.2.} {\it (i)
The key conditions C4 and C5, and the conclusion in the corollary show that if the artificially established relationship $\bm\beta=\bm s(\bm\alpha)$ has a certain sensibility, the correlation among covariates in $Q$ is weak, and sample size of $P$ is larger than that of $Q$, then, the asymptotic variance of the transfer learning estimator is of the order $1/n$.
This ensures that the transfer learning estimator has the convergence rate of the order $\sqrt n$.
It also implies the similar phenomenon of Stein's paradox as well, i.e., even if the source model is unrelated to the target model, the information of the source model can help to improve the estimation accuracy of the parameters in the target model.

(ii) The condition C5 implies that if the correlation between $X^Q$ and $Z^Q$ is relatively weak, then the inference on $Q$ can be improved by the information of $P$. Moreover, when $\tau\neq 0$, the convergence rate in probability is of the order $\sqrt{n}=\sqrt{cn_Q}$ for some constant $c>1$, which cannot be immoderately enhanced via increasing the sample size $n_P$; otherwise the convergence rate can be significantly enhanced via increasing the sample size $n_P$.

(iii)
Denote by $\widehat{\bm\alpha}^Q=((\widehat{\bm\gamma}^Q)^T,(\widehat{\bm\theta}^Q)^T)^T$ the maximum likelihood estimator of $\bm\alpha=(\bm\gamma^T,\bm\theta^T)^T$ by the likelihood function defined in model $Q$ as
$$l_Q(\bm\alpha)=\sum_{i=1}^{n_Q}\left(Y_i(\bm\gamma^TX^Q_i+
\bm\theta^T Z^Q_i) -G_Q(\bm\gamma^TX^Q_i+
\bm\theta^T Z^Q_i)\right).$$ Similar to Theorem 3.3, the estimator has the following asymptotic normality:
$$\left(F^Q_{n_Q}(\bm\alpha^0)\right)^{1/2}(\widehat{\bm\alpha}^Q-\bm\alpha^0)\rightarrow_d N(0,I).$$ Note that $F_{n}(\bm\alpha^0)-F^Q_{n_Q}(\bm\alpha^0)$ is a nonnegative definite matrix. By comparing the asymptotic covariances of the two estimators $\widehat{\bm\gamma}$ and $\widehat{\bm\gamma}^Q$, we can see that by transferring the knowledge of the source model $P$ to the target model $Q$, usually, the improvement of the estimator of $\bm\gamma$ is significant in the sense of convergence rate.

(iv) However, this is not always the case.  Note that the method above is derived from the likelihood equations (2.2) and (2.4), the likelihood equations respectively for $\bm\beta$ and $\bm\gamma$, but not for $\bm\theta$. Then, the transfer learning method cannot improve the estimator of $\bm\theta$. More generally, it follows from Theorem 3.3 that $$\frac{1}{(\bm\lambda^T F_n^{-1}(\bm\alpha^0)\bm\lambda)^{1/2}}\bm\lambda^T(\widehat{\bm\alpha}-\bm\alpha^0)\rightarrow_d N(0,I),$$ where nonzero vector $\bm\lambda\in \mathbb{R}^{d_1+d_2}$. Consider the setting where the direction $\bm\lambda$ satisfies $\bm\lambda^T F_n^{-1}(\bm\alpha^0)\bm\lambda=(\bm\xi^T F_n(\bm\alpha^0)\bm\xi)^{-1}$ for a nonzero vector $\bm\xi$ satisfying $\bm\xi^T \dot{\bm s}^T(\bm\alpha^0)=0$. It can be verified that such directions $\bm\xi$ and $\bm\lambda$ can be easily obtained, because the row vectors of the matrix $d_1\times(d_1+d_2)$-matrix $\dot{\bm s}(\bm\alpha^0)$ are always linearly correlative. Then, in the direction $\bm\lambda$, the estimator $\bm\lambda^T\widehat{\bm\alpha}$ only has the same asymptotic behavior as that of $\bm\lambda^T\widehat{\bm\alpha}^Q$, implying that the source model $P$ is not informative for such a parameter $\bm\lambda^T{\bm\alpha}$ in target model $Q$. As shown above, it is because that the likelihood is not for $\bm\theta$.}

\setcounter{equation}{0}
\section{Extension to generalized partially linear models}

The proposed strategy in the previous section reveals that the parametric information in source models with the structure of GLM can be easily transferred to a general target model, regardless of the complexity of the target model. Thus, we consider the scenario where the target model is relatively complex.
Specifically, here the source model $P$ is still a GLM as (2.1), but the target model $Q$ is a semiparametric model. More general extensions will be given in the Appendix. In current section, model $Q$ is chosen as the following generalized partially linear model:
\begin{eqnarray}
E[Y^Q|X^Q,Z^Q]=
g_Q^{-1}(\bm\gamma^TX^Q+t(Z^Q)),\end{eqnarray} where $\bm\gamma$ is a parametric vector with dimension $d_1$ and $t(\cdot)$ is an unknown nonparametric function. Here we suppose $Z^Q$ is a real-valued variable for the simplicity of notation.
Under an exponential family distribution, the log-likelihood of model $Q$ has the form:
\begin{eqnarray}
l_Q(\bm\gamma,t|Y^Q,X^Q,Z^Q)\propto Y^Q\left(\bm\gamma^TX^Q+t(Z^Q)\right) -G_Q(\bm\gamma^TX^Q+t(Z^Q)).\end{eqnarray}
 Theoretically, the true value of $\bm\gamma$ is defined as the solution to the following likelihood equation:
\begin{eqnarray}
 E[Y^QX^Q-  X^Q\dot{G}_Q(\bm\gamma^TX^Q+t(Z^Q))]=0.\end{eqnarray}
Similar to Lemma 2.1, by likelihood equation (4.3) together with likelihood equation (2.2), we have the following accurate representation:
\begin{eqnarray}
E[X^P \dot{G}_P(\bm\beta X^P)]=\Lambda  E[X^Q  \dot{G}_Q(\bm\gamma^TX^Q+t(Z^Q))],\end{eqnarray} where $\Lambda$ is the same as in the previous section, and it has been estimated by historical data or historical characteristics of $(X^P,Y^P)$ and $(X^Q,Y^Q)$. Actually, similar to (2.7) in Remark 2.1, the above builds an accurate relationship between the parameters in the two models.

Consider the special regime where model $P$ is a linear model but model $Q$ is a partially linear model as in (4.1). Similar to Theorem 2.2, by representation (4.4), we get an explicit expression for $\bm\beta$ as
\begin{eqnarray}
\bm\beta =(E[X^P(X^P)^T])^{-1}\Lambda E[X^Q \dot{G}_Q(\bm\gamma^TX^Q+t(Z^Q))].\end{eqnarray} In the above, $\bm\beta$ is a non-linear function of $\bm\gamma$ and $t(\cdot)$.

Generally, suppose that model $P$ is a general GLM as in (2.1) but $X^P\sim N(0,I_{d_1})$, similar to Theorem 2.3, we have the explicit expression for $\bm\beta$ as
\begin{eqnarray}
\bm\beta =(E\{Var[Y^P|X^P])^{-1}\Lambda E[X^Q\dot{G}_Q(\bm\gamma^TX^Q+t(Z^Q))].\end{eqnarray} Here $\bm\beta$ is a non-linear function of $\bm\gamma$ and $t(\cdot)$ as well.
In the representations (4.5) and  (4.6), we use the same correlation-ratio matrix $\Lambda$ to combine the parameters in the two models.

We can use the method of profile likelihood to deal with the above semiparametric problem (see, e.g., Severini and Wong, 1992; Severini and Staniswalis, 1994). First, for a given $\bm\gamma$, the local form of likelihood function (4.2) can be expressed as
\begin{eqnarray}\label{(partiallinear likelihood)}l_{Q}(t|\bm\gamma)= \sum_{i=1}^{n_Q}\left(Y^Q_i\left(\bm\gamma^TX^Q_i+t(Z^Q)\right)-  {G}_Q(\bm\gamma^TX^Q_i+t(Z^Q))\right)K_h((Z^Q_i-Z^Q)),
 \end{eqnarray} where $K_h(\cdot)=h^{-1}K(\cdot/h)$, $K(\cdot)$
is a kernel function and $h$ is the bandwidth. The estimator of $t(\cdot)$ is then obtained as \begin{eqnarray}\label{(t-partiallinear-estimator)}
\widehat{t}_{\bm\gamma}(Z^Q)
= \arg\max_{t\in \cal T} l_{Q}(t|\bm\gamma),\end{eqnarray} where $\cal T$ is a function space. Actually, under some regularity conditions, $\widehat{t}_{\bm\gamma}(Z^Q)$ is a consistent estimator of the function: $\bm\gamma\in \Gamma\rightarrow t_{\bm\gamma}(Z^Q)\in \mathbb{R}$, where
$$t_{\bm\gamma}(Z^Q)=\arg\max_{t\in \cal T}E\left[Y^Q\left(\bm\gamma^TX^Q+t(Z^Q)\right)-  {G}_Q(\bm\gamma^TX^Q+t(Z^Q))\right].$$ The corresponding curve $(\bm\gamma, t_{\bm\gamma})$ for $\bm\gamma\in \Gamma$ and $ t_{\bm\gamma}(Z^Q)\in \mathbb{R}$ is named as least favorable curve, with which the semiparametric efficiency can be achieved (see, e.g.,  Severini and Wong, 1992).

Then, by plugging the estimator $\widehat{t}_{\bm\gamma}(Z^Q)$ into (4.5) and (4.6), we get the corresponding explicit expressions of $\bm\beta$ that depend only on $\bm\gamma$ as
\begin{eqnarray}
&&\bm\beta =(E[X^P(X^P)^T])^{-1}\Lambda E[X^Q \dot{G}_Q(\bm\gamma^TX^Q+\widehat t_{\bm\gamma}(Z^Q))],\\ &&
\bm\beta =(E\{Var[Y^P|X^P])^{-1}\Lambda E[X^Q\dot{G}_Q(\bm\gamma^TX^Q+\widehat t_{\bm\gamma}(Z^Q))].\end{eqnarray}
The explicit representations in (4.9) and (4.10), and their empirical versions can be unified expressed respectively as
\begin{eqnarray*}\bm\beta =\bm s(\bm\gamma)=\widehat{\bm s}(\bm\gamma)+o_p(1)\ \mbox{ uniformly for } \bm\gamma\in \Gamma.\end{eqnarray*}
As a result, the transfer learning likelihood estimator $\widehat{\bm\gamma}$ can be obtained by maximizing the corresponding cr-TLL, namely,
\begin{eqnarray}\label{(partiallinear-estimator)}
\widehat{\bm\gamma}
= \arg\max_{\bm\gamma\in \Gamma} \widehat l_{(P,Q)}(\bm\gamma,\widehat t_{\bm\gamma}\,),\end{eqnarray} where the transfer learning likelihood function is \begin{eqnarray}\label{(gamma-partiallinear likelihood)}\nonumber\hspace{-4ex}\widehat l_{(P,Q)}(\bm\gamma,\widehat t_{\bm\gamma}\,) &=&\widehat w_P\sum_{i=1}^{n_P}\left(Y^P_i{\widehat {\bm s}}^T(\bm\gamma)X^P_i
 -G(\widehat {\bm s}^T(\bm\gamma)X^P_i )\right)\\&& \nonumber +\widehat w_Q\sum_{i=1}^{n_Q}\left(Y^Q_i\left(\bm\gamma^TX^Q_i+
 \widehat{t}_{\bm\gamma}(Z^Q_{-i})\right)-  {G}_Q(\bm\gamma^TX^Q_i+\widehat{t}_{\bm\gamma}(Z^Q_{-i}))\right),
 \end{eqnarray} in which $\widehat{t}_{\bm\gamma}(Z^Q_{-i})$ is the leave-one-out version of $\widehat{t}_{\bm\gamma}(Z^Q)$ valued at $Z^Q_i$. Finally, the estimator of nonparametric function is $\widehat{t}_{\widehat{\bm\gamma}}(Z^Q)$.

Under some regularity conditions (see, e.g.,  H\"{a}rdle et al., 2004), the theoretical properties of $\widehat{\bm\gamma}$ and $\widehat{t}_{\widehat{\bm\gamma}}(Z^Q)$ are similar to those in the previous section. For example, with a significant correlation-ratio, the estimator $\widehat{\bm\gamma}$ can achieve the global convergence rate of order $O_p(1/\sqrt{n})$, and the estimator $\widehat{t}_{\widehat{\bm\gamma}}(Z^Q)$ can achieve the standard nonparametric rate which depends on the smoothness of $t(\cdot)$ and the choice of kernel function $K(\cdot)$; the details are omitted here.

\setcounter{equation}{0}
\section{Numerical studies}
\subsection{Empirical evidences}
In this subsection, we provide the main results of simulation studies for the case when source models are not similar to the target model. The simulation studies for the case with similarity condition will be given in the Appendix. We examine our approach under the following two typical models: linear models and (log-link) Poisson models, with fixed sample size $n_Q$ and varying sample size $n_P$, and the opposite case.
The estimation performance is measured with the mean square error (mse) derived by  1000 replications. Moreover, the mean and the standard deviation (sd) of the simulations are also reported to show the special features of our method. For a comprehensive comparison, we consider the following competing estimators:
\begin{enumerate}
  \item The maximum likelihood estimation (MLE) obtained by the data only from the target model;
  \item The debiased method (DME) by the transfer learning method in  Tian and Feng (2021).
\end{enumerate}

\subsubsection{Linear Models}
We first consider a simple case where source model $P$ and target model $Q$ are the following linear models:
\begin{align*}
  \text{Model }P: ~~&Y^P=\beta^T X^P+\varepsilon^P,\\
  \text{Model }Q: ~~&Y^Q=\gamma^T X^Q+\theta^T Z^Q+\varepsilon^Q
\end{align*}
with $\varepsilon^P\sim N(0,1)$ and $\varepsilon^Q\sim N(0,1)$. Here the covariates $X^P=(X^{1P}, X^{2P}, X^{3P})^T\sim N(\bm{0},\Sigma_3)$ and $((X^Q)^T,(Z^Q)^T)^T=(X^{1Q}, X^{2Q}, X^{3Q}, Z^{1Q}, Z^{2Q})^T\sim N(\bm{0},\Sigma_5)$, where $\Sigma_d=(\sigma_{ij})_{1\leq i,j\leq d}$ with $\sigma_{ij}=0.2^{|i-j|}$. The model parameters are set as $\beta=(1,1,1)^T$, $\gamma=(1,0.8,-1)^T$ and $\theta=(1,-1)^T$. It can be seen that the difference between the regression coefficients in the models is highly significant, implying that the similarity condition is violated completely.

The correlation-ratio matrix $\Lambda$ is estimated by the historical data of the variables $(X^P,Y^P)$ and $(X^Q,Y^Q)$. The simulation results are reported in Table 1. We have the following findings:
\begin{enumerate}
  \item Our estimator of cr-TLL is much better than DME and MLE under the linear models in the sense that the mean square error of cr-TLL estimation is significantly smaller than those of DME and MLE.
  \item When the data set size $n_Q$ of the target model $Q$ is fixed, the standard deviation of cr-TLL estimation decreases fast with the increase in $n_P$, while the bias of cr-TLL estimation increases slightly (see Part I of Table 1).
  \item When $n_P=1000$ is fixed, both the bias and the standard deviation of cr-TLL estimation decrease with the increase in $n_Q$ (see Part II of Table 1). It is worth noting that our  method performs very well for small sample data of model $Q$. Particularly, when $n_Q=10$ the mse of MLE is about 0.3, but ours is only 0.0003.
\end{enumerate}

\begin{sidewaystable}[thp]  \scriptsize
\caption{\label{tab:1} The mean, sd and mse of the cr-TLL estimator, DME and MLE under the linear models.1}
\centering
\begin{tabular}{|c|c|ccc|ccc|ccc|ccc|ccc| }
\hline
\multicolumn{17}{|c|}{Part I: fixed $n_Q=50$ and varying $n_P$}\\
\hline
\multicolumn{2}{|c|}{$n_P$}&\multicolumn{3}{|c|}{$50$}&\multicolumn{3}{|c|}{$100$}&\multicolumn{3}{|c|}{$200$}&
\multicolumn{3}{|c|}{$500$}&\multicolumn{3}{|c|}{$1000$}\\
\multicolumn{2}{|c|}{}& cr-TLL & DME&MLE & cr-TLL & DME&MLE & cr-TLL & DME&MLE & cr-TLL & DME&MLE & cr-TLL&DME&MLE \\
\hline
\multirow{5}{*}{mean}&$\gamma_1$&0.9752&1.0016&1.0015&0.9705&1.0004&1.0015&0.9710&1.0018&1.0015&0.9713& 1.0023&1.0015&0.9705&0.9999&1.0015\\
&$\gamma_2$&0.8515& 0.7989&	0.7992&0.8569& 0.7998&	0.7992&0.8576&0.7936&0.7992& 	0.8573&0.8013&0.7992&0.8585& 0.7984&0.7992\\
&$\gamma_3$&-1.0382& -1.0028&-1.0031 &-1.0407 &	-1.0028	&-1.0031&-1.0407& -1.0026& -1.0031&-1.0407 &-1.0041&-1.0031 &	-1.0411 &-1.0009	&-1.0031 \\
&$\theta_1$&1.0247&1.0026&1.0028&1.0273& 1.0052	&1.0028& 1.0313 &1.0033	&1.0028& 	1.0335 &1.0034&1.0028 &		1.03417 &1.0017	&1.0028 \\
&$\theta_2$ &-0.9898&-0.9972&-0.9980 &-0.9911& -0.9962&-0.9980 &-0.9911 &-0.9974		&-0.9980 &-0.9901 &-0.9968&-0.9980 &-0.9908 &-0.9992&-0.9980\\
\hline
\multirow{5}{*}{sd}&$\gamma_1$&0.0952&0.1539&0.1535& 0.0724 & 0.1553&0.1535 &	0.0539 &0.1548&0.1535 &	0.0347 &0.1551&0.1535 	&0.0253&0.1571 	&0.1535\\
&$\gamma_2$&0.0519 &0.1570&0.1561 &	0.0373 &0.1608&0.1561 &	0.0260 &0.1590&0.1561& 	0.0162 &0.1821&0.1561&0.0123 &0.1570&0.1561\\
&$\gamma_3$&0.0440 &0.1572&0.1551 &	0.0362 &0.1574&0.1551& 	0.0318&0.1585&0.1551& 	0.0283 &0.1578&0.1551 	&	0.0265 	&0.1621&0.1551\\
&$\theta_1$&0.0802 &0.1606	&0.1593 &0.0596 &0.1799	&0.1593 &0.0479 &0.1601	&0.1593 &	0.0372 &0.1623&0.1593 &	0.0326 &0.1671&0.1593\\
&$\theta_2$&0.0501 &0.1538&0.1516 &0.0336& 0.1540&0.1516 &	0.0244 &0.1530	&0.1516& 	0.0152 &0.1549&0.1516 &	0.0111 &0.1560&0.1516\\
\hline
\multirow{5}{*}{mse}&$\gamma_1$&0.0097&0.0236&0.0235&0.0061& 0.0241&0.0235&	0.0037& 0.0239&0.0235 &	0.0020 &0.0240&0.0235 &	0.0015&0.0246 	&0.0235\\
&$\gamma_2$&0.0053 &0.0246&0.0243 &	0.0046 &0.0258&0.0243 &	0.0040 &0.0252&0.0243& 	0.0035&0.0331&0.0243 	&0.0036&0.0246 	&0.0243\\
&$\gamma_3$&0.0034& 0.0247&0.0241 &0.0030 &0.0247&0.0241 &	0.0027 &0.0251&0.0241 &	0.0025&0.0249&0.0241 &	0.0024 &0.0262		&0.0241\\
&$\theta_1$&0.0070& 0.0257&0.0254&0.0043& 0.0323&0.0254&0.0033&0.0256&0.0254&0.0025 &		0.0263&0.0254&0.0022 &0.0279		&0.0254\\
&$\theta_2$&0.0026 &0.0236&0.0230 &	0.0012 &0.0237&0.0230 &	0.0007 &0.0233&0.0230& 	0.0003 &0.0240&0.0230& 	0.0002& 0.0243	&0.0230\\
\hline
\hline
\multicolumn{17}{|c|}{Part II: fixed $n_P=1000$ and varying $n_Q$}\\
\hline
\multicolumn{2}{|c|}{$n_Q$}&\multicolumn{3}{|c|}{$10$}&\multicolumn{3}{|c|}{$50$}&\multicolumn{3}{|c|}{$100$}&
\multicolumn{3}{|c|}{$200$}&\multicolumn{3}{|c|}{$500$}\\
\multicolumn{2}{|c|}{}& cr-TLL & DME&MLE & cr-TLL & DME&MLE & cr-TLL & DME&MLE & cr-TLL & DME&MLE & cr-TLL&DME&MLE \\
\hline
\multirow{5}{*}{mean}&$\gamma_1$&1.0531&1.0227&1.0193&1.0517&0.9985&0.9979&1.0507& 1.0026 &1.0017&1.0484&1.0021&1.0017&1.0428&0.9985  &0.9986\\
&$\gamma_2$&0.7810 &0.7997 &0.7973 &0.7808 &0.8092 &0.8092 &0.7813 &0.8027 &0.8002 &0.7819 &0.7932 &0.7937 &0.7841 &0.7998  &0.8001\\
&$\gamma_3$&-1.0306&-0.9818&-0.9817&-1.0251&-0.9965&-0.9964&-1.0260&-1.0002&-1.0007&-1.0255& -1.0034&-1.0033&-1.0256&-1.0014  &-1.0014 \\
&$\theta_1$&0.9223 &1.0208 &1.0224 &0.9230 &0.9981 &0.9972 &0.9239 &1.0010 &1.0018 &0.9253 &0.9976&0.9974 &0.9273 &1.0003  &1.0012\\
&$\theta_2$&-1.0135&-0.9936&-0.9935&-1.0135&-0.9980&-0.9982&-1.0133&-1.0000&-1.0012&-1.0128& -0.9976&-0.9979&-1.0117&-0.9983  &-0.9991\\
\hline
\multirow{5}{*}{sd}&$\gamma_1$&0.0247&0.4934&0.4839&0.0230&0.1565&0.1538&0.0225&0.1117  &0.1064&0.0221&0.0755&0.0723&0.0210&0.0495  &0.0452\\
&$\gamma_2$&0.0159&0.5219&0.5100&0.0112&0.1572&0.1561&0.0109&0.1122&0.1065&0.0106&0.0749  &0.0735&0.0104&0.0505&0.0472\\
&$\gamma_3$&0.0789&0.5246&0.5168&0.0298&0.1583&0.1568&0.0213&0.1094&0.1053&0.0156&0.0763  &0.0752&0.0107&0.0529&0.0468\\
&$\theta_1$&0.0794&0.5349&0.5302&0.0321&0.1592&0.1569&0.0236&0.1095&0.1068&0.0179&0.0820  &0.0754&0.0148&0.0488&0.0460\\
&$\theta_2$&0.0145&0.5261&0.5211&0.0114&0.1496&0.1474&0.0110&0.1098&0.1044&0.0105&0.0742  &0.0724&0.0106&0.0492&0.0472\\
\hline
\multirow{5}{*}{mse}&$\gamma_1$&0.0034&0.2437&0.2343&0.0032&0.0244&0.0236&0.0030&0.0124&0.0113&0.0028&0.0057  &0.0052&0.0022&0.0024  &0.0020\\
&$\gamma_2$&0.0006&0.2721&0.2599&0.0004&0.0247&0.0244&0.0004&0.0126&0.0113&0.0004&0.0056  &0.0054&0.0003&0.0025  &0.0022\\
&$\gamma_3$&0.0071&0.2753&0.2672&0.0015&0.0250&0.0246&0.0011&0.0119&0.0110&0.0008&0.0058  &0.0056&0.0007&0.0028  &0.0021\\
&$\theta_1$&0.0123&0.2863&0.2813&0.0069&0.0253&0.0246&0.0063&0.0119&0.0114&0.0059&0.0067  &0.0057&0.0054&0.0023  &0.0021\\
&$\theta_2$&0.0003&0.2765&0.2713&0.0003&0.0223&0.0217&0.0002&0.0120&0.0109&0.0002&0.0055 &0.0052&0.0002&0.0024  &0.0022\\
\hline
\end{tabular}
\end{sidewaystable}

\subsubsection{Poission Model}
The log-link Poission regression functions are chosen as
\begin{align*}
  \text{Model }P: ~~&\ln(E[Y^P|X^P])=\beta^T X^P,\\
  \text{Model }Q: ~~&\ln(E[Y^Q|X^Q,Z^Q])=\gamma^T X^Q+\theta^T Z^Q
\end{align*}
with $\beta=(1,1,1)^T$, $\gamma=(1,0.8,-1)^T$ and $\theta=(1,-1)^T$. Also the two models are significantly different, the similarity condition being violated utterly. The conditional distributions of $Y^P$ and $Y^Q$ given the corresponding covariates are Poisson of mean $E[Y^P|X^P]$ and $E[Y^Q|X^Q,Z^Q]$, respectively. The independent samples of explanatory variables $X^P$, $X^Q$ and $Z^Q$ are generated from multivariate normal law with the mean equal to 0, the standard deviations equal to 1 and the correlation equal to 0. The correlation ratio matrix $\Lambda$ is estimated by the historical data of the variables $(X^P,Y^P)$ and $(X^Q,Y^Q)$.

The simulation results are reported in Table 2. We have the following
findings that are similar to the results in linear models:
\begin{enumerate}
  \item Our method of cr-TLL is much better than DME and MLE in the poisson models under the criterion of mse.
  \item When the data set size $n_Q$ of the target model $Q$ is fixed, the standard deviation of cr-TLL estimation decreases fast with the increase in $n_P$, while the bias of cr-TLL estimation increases insignificantly (see Part I of Table 2).
  \item When $n_P=1000$ is fixed, both the bias and the standard deviation of cr-TLL estimation decrease with the increase in $n_Q$ (see Part II of Table 2), and our method still works very well for the case when the sample size of model $Q$ is small.

\end{enumerate}

In short, the simulations results above (together with the simulations given in the Appendix) clearly illustrate that our method is better than the competitors for all the models with or without the similarity condition.

\begin{sidewaystable}[thp]  \scriptsize
\caption{\label{tab:3} The mean, sd and mse of the cr-TLL estimator, DME and MLE under the Poisson models with fixed $n_Q=50$ and varying $n_P$, and the known distributions of covariates. }
\centering
\begin{tabular}{|c|c|ccc|ccc|ccc|ccc| }
\hline
\multicolumn{14}{|c|}{Part I: fixed $n_Q=50$ and varying $n_P$}\\
\hline
\multicolumn{2}{|c|}{$n_P$}&\multicolumn{3}{|c|}{$100$}&\multicolumn{3}{|c|}{$200$}&
\multicolumn{3}{|c|}{$500$}&\multicolumn{3}{|c|}{$1000$}\\
\multicolumn{2}{|c|}{}& cr-TLL & DME&MLE & cr-TLL & DME&MLE & cr-TLL & DME&MLE & cr-TLL & DME&MLE \\
\hline
\multirow{5}{*}{mean}&$\gamma_1$&0.9727&1.0022&0.9974&0.9675&1.0025&0.9974&0.9622&1.0022&0.9974&0.9587&0.9990&0.9974\\
&$\gamma_2$&0.8134&0.7997&0.8016&0.8209&0.7992&0.8016&0.8274&0.7989&0.8016&0.8292&0.8045&0.8016\\
&$\gamma_3$&-0.9883&-0.9847&-0.9875&-0.9875&-0.9844&-0.9875&-0.9871&-0.9823&-0.9875&-0.9858&-0.9830&-0.9875\\
&$\theta_1$&0.9111&0.9987&1.0007&0.8914&0.9941&1.0007&0.8706&0.9992&1.0007&0.8593&0.9987&1.0007\\
&$\theta_2$&-1.0010&-0.9880&-0.9968&-1.0048&-0.9877&-0.9968&-1.0125&-0.9900&-0.9968&-1.0148&-0.9880&-0.9968\\
\hline
\multirow{5}{*}{sd}&$\gamma_1$&0.0412&0.1223&0.1124&0.0327&0.1235&0.1124&0.0243&0.1340&0.1124&0.0289&0.1506&0.1124\\
&$\gamma_2$&0.0435&0.1109&0.1060&0.0345&0.1375&0.1060&0.0230&0.1118&0.1060&0.0269&0.1297&0.1060\\
&$\gamma_3$&0.0580&0.1174&0.1095&0.0476&0.1176&0.1095&0.0454&0.1279&0.1095&0.0516&0.1217&0.1095\\
&$\theta_1$&0.0505&0.1149&0.1101&0.0440&0.1280&0.1101&0.0462&0.1198&0.1101&0.0347&0.1407&0.1101\\
&$\theta_2$&0.0564&0.1245&0.0980&0.0556&0.1296&0.0980&0.0437&0.1342&0.0980&0.0415&0.1501&0.0980\\
\hline
\multirow{5}{*}{mse}&$\gamma_1$&0.0024&0.0149&0.0126&0.0021&0.0152&0.0126&0.0020&0.0179&0.0126&0.0025&0.0226&0.0126\\
&$\gamma_2$&0.0020&0.0123&0.0112&0.0016&0.0189&0.0112&0.0012&0.0125&0.0112&0.0015&0.0168&0.0112\\
&$\gamma_3$&0.0034&0.0140&0.0121&0.0024&0.0140&0.0121&0.0022&0.0166&0.0121&0.0028&0.0150&0.0121\\
&$\theta_1$&0.0104&0.0131&0.0121&0.0137&0.0164&0.0121&0.0188&0.0143&0.0121&0.0209&0.0197&0.0121\\
&$\theta_2$&0.0031&0.0156&0.0096&0.0031&0.0169&0.0096&0.0020&0.0180&0.0096&0.0019&0.0226&0.0096\\
\hline
\hline
\multicolumn{14}{|c|}{Part II: fixed $n_P=1000$ and varying $n_Q$}\\
\hline
\multicolumn{2}{|c|}{$n_Q$}&\multicolumn{3}{|c|}{$50$}&\multicolumn{3}{|c|}{$100$}&
\multicolumn{3}{|c|}{$200$}&\multicolumn{3}{|c|}{$500$}\\
\multicolumn{2}{|c|}{}& cr-TLL & DME&MLE & cr-TLL & DME&MLE & cr-TLL & DME&MLE & cr-TLL& DME&MLE \\
\hline
\multirow{5}{*}{mean}&$\gamma_1$&0.9990&1.0010&0.9974&0.9998&0.9987&0.9977&1.0000&1.0020&1.0030&1.0001&1.0002&1.0002\\
&$\gamma_2$&0.7690&0.8011&0.8032&0.7706&0.7984&0.7973&0.7725&0.7976&0.7942&0.7781&0.8012&0.8001\\
&$\gamma_3$&-1.0262&-0.9840&-0.9889&-1.0275&-0.9985&-0.9978&-1.0235&-0.9964&-0.9969&-1.0175&-0.9991&-1.0002\\
&$\theta_1$&1.0130&0.9957&0.9933&1.0131&0.9997&1.0008&1.0152&0.9994&0.9993&1.0161&1.0016&0.9994\\
&$\theta_2$&-1.0063&-0.9918&-0.9916&-1.0069&-0.9951&-0.9972&-1.0080&-0.9978&-0.9994&-1.0060&-0.9973&-0.9991\\
\hline
\multirow{5}{*}{sd}&$\gamma_1$&0.0228&0.1037&0.1080&0.0166&0.0573&0.0709&0.0142&0.0362&0.0542&0.0097&0.0232&0.0207\\
&$\gamma_2$&0.0201&0.1036&0.1165&0.0164&0.0578&0.0655&0.0147&0.0388&0.0587&0.0118&0.0280&0.0218\\
&$\gamma_3$&0.0615&0.1047&0.1127&0.0200&0.0554&0.0652&0.0187&0.0427&0.0573&0.0137&0.0312&0.0215\\
&$\theta_1$&0.0257&0.0907&0.1020&0.0177&0.0577&0.0706&0.0138&0.0397&0.0453&0.0110&0.0268&0.0242\\
&$\theta_2$&0.0524&0.0958&0.1175&0.0216&0.0590&0.0662&0.0184&0.0439&0.0511&0.0134&0.0282&0.0229\\
\hline
\multirow{5}{*}{mse}&$\gamma_1$&0.0005&0.0107&0.0116&0.0002&0.0032&0.0050&0.0002&0.0013&0.0029&0.0000&0.0005&0.0004\\
&$\gamma_2$&0.0013&0.0107&0.0135&0.0011&0.0033&0.0043&0.0009&0.0015&0.0034&0.0006&0.0007&0.0004\\
&$\gamma_3$&0.0044&0.0112&0.0128&0.0011&0.0030&0.0042&0.0009&0.0018&0.0032&0.0004&0.0009&0.0004\\
&$\theta_1$&0.0008&.00824&0.0104&0.0004&0.0033&0.0049&0.0004&0.0015&0.0020&0.0003&0.0007&0.0005\\
&$\theta_2$&0.0027&0.0092&0.0138&0.0005&0.0035&0.0043&0.0004&0.0019&0.0026&0.0002&0.0008&0.0005\\
\hline
\end{tabular}
\end{sidewaystable}

\subsection{Real data analysis for airline delays}

To illustrate the usefulness of our proposed transfer learning method in practice, we examine the on-time performance of domestic flights by U.S. air carriers. The
2009 ASA Data Expo\footnote{http://stat-computing.org/dataexpo/2009/the-data.html} (see  DVN, 2008)  consists of flight departure/arrival information for all commercial flights within the United States, from October 1987 to April 2008, with $n = 123,534,969$ observations. The raw data set can separate into two parts:
\begin{description}
  \item[Part I:] Observations from October 1987 to May 2003 only contain 24 basic variables, such as the scheduled departure and arrival times, the actual departure and arrival times, the origin and destination airports, the arrival delay and departure delay time, and so on.
  \item[Part II:] Observations from June 2003 to  April 2008 contain 24 basic variables and 5 causal variables of flight delay recorded in minutes (the carrier delay time, the weather delay time, the national aviation system delay time, the security delay time and the late air craft delay time).
\end{description}
We regard Part I as the data set from the source model $P$ and Part II as the data set from the target model $Q$. In our analysis, the arrival delay time  is referred as to the response variable, while the departure delay time (Departure) and  the flight distance (Distance) are thought of as the covariates in model $P$, and Departure, Distance, the carrier delay time (Carrier), the weather delay time (Weather), the national aviation system delay time (NAS), the security delay time (Security) and the late air craft delay time (LastAircraft) are designed as the explanatory variables in model $Q$.

According to the feature of data sets, we consider a logistic regression with the Bernoulli distribution to describe whether or not the arrival delay is equal to or longer than 15 min. More specifically, in the logistic regression model, the response variable is equal to 0 if a flight is late by less than 15 min, and 1 otherwise. We utilize the data of May 30 and 31, 2003 as the sample from Model $P$ and the data of the first half of June 1, 2003 as the sample from Model $Q$ to predict the arrival delay of rest flights on June 1, 2003. The historical data used to estimate $\Lambda$ are chosen as the data of May 29, 2003 from Molde $P$ together with the data of the first half of June 1, 2003 from Model $Q$. Our method is compared with the maximum likelihood estimation (MLE) which is estimated with the data of the first half of June 1, 2003. We delete the incomplete observations and standardize the data before modeling. The estimates of parameters and the component wise confidence intervals are given in Table 3. As can be seen from Table 3, our result is more reasonable. The reasons are listed as follows:
\begin{enumerate}
  \item According to the perception of flight delay, the covariate of departure delay and 5 causal variables of flight delay should be the main factors that cause flight delay. By our method, the estimated regression coefficients of these covariates are positive, which is consistent with the perception of flight delay. However, the come estimated regression coefficients by MLE are not positive.
  \item The coefficient estimate of the flight distance is negative, this means that the further a plane flies, the less chance of delays. It is the same as our understanding. On June 1, 2003, the delay rate of the flight whose flight distance is longer than 2000 km is $26.88\%$ while it is $30.75\%$ when the flight distance is less than 2000 km.
  \item For most regression coefficients, the length of bootstrap confidence intervals by our method are shorter than those by MLE, meaning that our method is more accurate.
\end{enumerate}
From the point of view of prediction, our method only produces 81 wrong predicting outcomes, which is much smaller than 9573, the number of wrong predicting outcomes by MLE. There are totally 10119 flights in the second half of June 1, 2003. Thus, our method obtains a very high prediction accuracy, $99.20\%$.

\begin{table}[htbp]   \footnotesize
\centering
\caption{\small The estimator and  the confidence intervals of the parameters for the 2009 ASA Data Expo: logistic regression model. \newline}
\label{tab:8}
\begin{tabular}{|c|cc|cc|}
\hline
\multirow{2}{*}{Variable}& \multicolumn{2}{|c|}{Coefficient estimate} &\multicolumn{2}{|c|}{ 0.9 Confidence interval}\\
&cr-TLL&MLE&cr-TLL&MLE\\
\hline
Departure & 0.1224&-1.0991 &(-7.7566, 12.2327)&(-3.6918, 13.1254)\\

Distance& -1.0521 &0.4628&(-1.7903, 0.4659)&(-0.3107, 3.2439)\\

Carrier&7.5356&0.6883&( -6.1150, 14.9264)&(-66.1801, 1.3611)\\

Weather&0.0928&11.5235&(-7.1883, 11.3431)&(-15.8101, 11.5137)\\

NAS&16.9073&-13.3688&(3.6990,  32.7233)&(-29.6694,  -2.3332)\\

Security&9.9734&7.2164  &(-7.4755,11.2615) &(-38.6473,11.9135) \\

LastAircraft&20.7729 &-0.04377  &(-0.4377,23.5689)&(-41.1963,0.3833)  \\
\hline
\end{tabular}
\end{table}

\setcounter{equation}{0}
\section{Conclusions and future works}

It was stated in Introduction that the existing transfer learning methodologies for parameter models need the similarity condition: the source model and target model share some parameters or priors, or the parameters in the source models are closed to those in the target model. With the condition, the existing methods can transfer knowledge across the tasks by encoding into the shared (or approximate) parameters or priors.
In the previous sections, we introduced a parameter-based transfer learning approach for the scenarios that contain both conditional distribution drift and covariate shift, without the classical similarity condition. Based on the special structure of generalized linear models, the relationship between the source models and target model was built by a newly proposed strategy: the correlation-ratios between the target model and source models. Then, the relationship among the parameters between the target model and source models was established by the precise representations. Based on these precise representations, the knowledge behind in the source models can be transferred to the target model to improve the convergence rate of parameter estimation, even to achieve the global convergence rate. The parameter estimation theories, including the asymptotic normality and the global convergence rate were established, and the method was extended into partially linear model and non-linear model.
The behavior of the transfer learning method was further illustrated by various numerical examples from simulation experiments and a real data analysis. The numerical examples verified that the
finite performance of the new method is much better than the competitors. Interestingly, the theoretical properties and numerical results illustrated the phenomenon of Stein's paradox, in other words, even for the case where the source models are unrelated to the target model, the information of source models can be used to improve the transfer learning estimation.

However, in the procedure of our correlation-ratio transfer learning, the correlation-ratio matrix $\Lambda$ is required to be known in advance or be estimated by historical data or historical characteristics. How to relax this condition is an important topic. On the other hand, in the previous sections we focused only on the exponential family with linear and partially linear links. It is still a challenge to extend the strategy to nonparametric models and general semiparametric models. These are
interesting issues and are worth further study in the future.

\section*{Acknowledgments}
The research of Lin Lu was supported by the National Key R\&D Program of China (grant No. 2018YFA0703900) and the National Natural Science Foundation of China(grant No. 11971265). The research of Li Weiyu was supported by the National Natural Science Foundation of China (grant No. 12001318).

\

\section*{References}
\begin{description}

\item Ben-David, S., Blitzer, J., Crammer, K., and Pereira, F. (2007). Analysis of representations for domain adaptation. {\it In Advances in Neural Information Processing Systems},
137-144.

\item Blitzer, J., Crammer, K., Kulesza, A., Pereira, F., and Wortman, J. (2008). Learning bounds for domain adaptation. {\it In Advances in Neural Information Processing Systems},
129-136.

\item Bonilla, E., Chai, K. M. and Williams, C. (2008). Multi-task Gaussian process prediction. {\it Proc. 20th Ann. Conf. Neural Information
Processing Systems},  153-160.

\item Cai, T. and Wei, H. (2021). Transfer learning for nonparametric classification: Minimax rate and  adaptive classifier. {\it Annals of Statistics}. 49(1):100-128.

\item Certo, S. T. (2003). Influencing initial public offering investors with prestige: Signaling with board structures. {\it Academy of Management Review}, {\bf 28}, 432-446.

\item Chikara, U. (2003). Sequential estimation of the ratio of two
exponential scale parameters. {\it Kyoto University Research Information Repository.} 84-98.

\item Choi, K., Fazekas, G., Sandler, M. and Cho, K. (2017). Transfer learning for music classification and regression tasks. arXiv preprint arXiv:1703.09179.

\item Data Expo 2009: Airline on time data, 2008. URL https://doi.org/10.7910/DVN/HG7NV7.

\item Desyllas, P. and Sako, M. (2013). Profiting from business model innovation: Evidence from pay-as-you-drive auto insurance. {\it Research Policy}, {\bf 42}, 101-116.

\item Evgeniou, T. and Pontil, M. (2004). Regularized multi-task learning. {\it Proc. 10th ACM SIGKDD Int'l Conf. Knowledge Discovery and Data
Mining}, 109-117.

\item Fahrmeir, L. and Kaufmann, H. (1985). Consistency and asymptotic normality of the maximum likelihood estimator in generalized linear models. {\it  Annals of Statistics},  {\bf 13}, 342-368.

\item Gao, J., Fan, W., Jiang, J. and Han, J. (2008). Knowledge transfer via multiple model local structure mapping. {\it Proc. 14th ACM SIGKDD Int¡¯l Conf. Knowledge Discovery and Data Mining},
 283-291.

\item Gong, B., Shi, Y., Sha, F., and Grauman, K. (2012). Geodesic
ow kernel for unsupervised domain adaptation. {\it 2012 IEEE Conference on Computer Vision and Pattern
Recognition}, 2066-2073.

\item Hansen, B. E. (2004). Nonparametric Conditional Density Estimation. https: scholar.google.com/scholar?hl=zh-CN\&as\_sdt=0\%2C14
    \&q=Nonparametric+ Conditional+Density+Estimation\&btnG=

\item H\"ardle, W., M\"uller, M., Sperlich, S. and Werwatz, A. (2004). {\it Nonparametric and Semiparametric models}. Springer, Berlin.

\item Henry, W. J., Reeve, Timothy, I., Cannings and
Richard,J. S. (2021). Adaptive transfer learning. {\it Annals of Statistics}, (to appear).

\item Hj\o rt, N. and D. Pollard (1993), Asymptotics for minimizers of convex processes.
{\it Statistical Research Report}. http://www.stat.rutgers.edu/home/ztan/material/hjort-pollard-convex.pdf.

\item Hood, L., Heath, J. R., Phelps, M. E., \& Lin, B. (2004). Systems biology and new technologies enable
predictive and preventative medicine. {\it Science}, {\bf 306}, 640-643.

\item Huang, J. T., Li, J., Yu, D., Deng, L., and Gong, Y. (2013). Cross-language knowledge transfer using multilingual deep neural network with shared hidden layers. {\it 2013 IEEE
International Conference on Acoustics, Speech and Signal Processing}, 7304-7308.

\item James, L. P., James, H. S. and Thomas, M. S. (1989). Semiparametric estimation of index coefficients. {\it Econometrica}, {\bf 57}, 1403-1430.

\item Mullen, K. (1967). A note on the ratio of two independent random variables. {\it Amer. Statist}, {\bf 21}, 30-31.

\item Kpotufe, S. and Martinet, G. (2021). Marginal singularity, and the benefits of labels in covariate-shift.  {\it Annals of Statistics}, {\bf 49(6)}:3299¨C3323.

\item Lawrence, N. D. and Platt, J. C. (2004). Learning to learn with the
informative vector machine. {\it Proc. 21st Int'l Conf. Machine
Learning}.

\item Lin, L., Lu, J.  and Li, W.Y. (2021). Online updating statistics for heterogenous updating regressions via homogenization techniques. https://arxiv.org/abs
    /2106.12370.

\item  Lin, L., Dong, P., Song, Y. and Zhu, L.X. (2017). Upper expectation parametric regression. {\it Statistica Sinica}, {\bf 27}, 1265-1280.

\item Mansour, Y., Mohri, M., and Rostamizadeh, A. (2009). Domain adaptation: Learning bounds and algorithms. {\it Proceedings of The 22nd Annual Conference on Learning
Theory (COLT 2009)}, Montr\'{e}al, Canada.

\item Pan, S. J. and Yang, Q. (2010). Survey on transfer learning. {\it IEEE Transactions of knowledge and data engineering}, {\bf 22}, 1345-1359.

\item Rupp, N. G. (2007). Further Investigations into the Causes of Flight Delays, Techcnical Report, Department
of Economy, East Carolina University.

\item Reeve H. W. J., Cannings, T. I. and Samworth, R. J. (2021). Adaptive transfer learning. {\it Annals of Statistics}. {\bf 49(6)}:3618 ¨C 3649.

\item Shimodaira, H. (2000). Improving predictive inference under covariate shift by weighting
the log-likelihood function. {\it Journal of statistical planning and inference}, {\bf 90}, 227-244.

\item Schwaighofer, A., Tresp, V. and Yu, K. (2005). Learning Gaussian process kernels via hierarchical Bayes. {\it Proc. 17th Ann. Conf. Neural
Information Processing Systems},  1209-1216.

\item Severini, T. A., and Staniswalis, J. G. (1994). Quasi-likelihood estimation in semiparametric models. {\it Journal of the
American Statistical Association}, {\bf 89}, 501-511.

\item Severini, T. A., and Wong, W. H. (1992). Profile likelihood and conditionally parametric models. {\it The Annals of
Statistics}, {\bf 20}, 1768-1802.

\item Stein, C. (1956). Inadmissibility of the Usual Estimator for the Mean of a
Multivariate Normal Distribution. in {\it Proceedings of the Third Berkeley
Symposium on Mathematical Statistics and Probability} (Vol. 1), Berkeley
and Los Angeles, University of California Press,  197-206.

\item Stein, C. (1986). {\it Approximate Computation of Expectations}.  IMS Lecture Notes-Monogr. Ser. Vol. 7, Inst.
Math. Statist.

\item Stigler, S. M. (1990). The 1988 Neyman Memorial Lecture: A Galtonian
Perspective on Shrinkage Estimators. {\it  Statistical Science}, {\bf 5}, 147-155.


\item Storkey, A. (2009). When training and test sets are different: characterizing learning transfer. {\it Dataset shift
in machine learning}, 3-28. MIT Press.

\item Sugiyama, M., Nakajima, S., Kashima, H., Buenau, P. V., and Kawanabe, M. (2008).
Direct importance estimation with model selection and its application to covariate shift
adaptation. {\it In Advances in Neural Information Processing Systems}, 1433-1440.

\item Tian, Y. and Feng, Y. (2021). Transfer Learning under High-dimensional Generalized Linear Models. https://arxiv.org/abs/2105.14328.

\item Tzeng, E., Hoffman, J., Saenko, K., and Darrell, T. (2017). Adversarial discriminative domain adaptation. {\it In Proceedings of the IEEE Conference on Computer Vision and Pattern Recognition}, 7167-7176.


\item Wang, C., Chen, M. H., Wu, J., Yan, J., Zhang, Y. and
Schifan, E. (2018). Online updating method with new variables for big data streams. {\it The Canadian Journal of Statistics}, {\bf 46}, 123-146.

\item Wei, B. C. (1998). {\it Exponential family nonlinear models}, Springer-Verlag Singapore Pte. Etd.

\item Weiss, K., Khoshgoftaar, T. M and Wang, D. D. (2016). A survey of transfer learning. {\it Big Data},  3:9, DOI 10.1186/s40537-016-0043-6.

\item Xia, R., Zong, C., Hu, X. and Cambria E. (2013). Feature ensemble plus sample selection: domain adaptation for sentiment classification.
{\it IEEE Intell Syst}, {\bf 28} (3):10-18.

\item Yang, Z., Balasubramanian, K., Wang, Z. and Liu, H. (2017)
Learning non-gaussian multi-index model via second-order stein¡¯s method. https://xueshu.baidu.com
/usercenter
/paper/show?paperid
=c0b406292f713d3513098993bfC744ca\&site
=xueshu\_se

\end{description}

\section{Appendix}

\subsection{Stein's paradox and the correlation to our model conditions}

In our paper, the issue is what follows: can the information of source model $P$ be transferred into the target model $Q$ to enhance the inferential accuracy of $Q$ if actually the two models are not similar? The issue can be classified as a variational Stein's paradox. Here we refer to the perspective of Stigler (1990) to briefly recall Stein's paradox.
A similar problem was asked in the existing literature:
can the information about the price of apples in Washington and about the price of oranges in Florida be used to improve an estimate of the price of French wine if the prices are unrelated to each other? When the phenomenon is first encountered it can seem preposterous. But the answer is YES!

According to Stigler (1990), we consider a simple situation: a collection of independent measurements $X_1, \cdots, X_k$ is available, each measuring a different $\theta_i$, and each normally distributed $N(\theta_i, 1)$. The ``ordinary" estimator of $\theta_i$ is $\widehat\theta^0_i=X_i$, but the estimator is inadmissible if $k\geq 3$, because James-Stein estimator
$$\widehat\theta^{JS}_i=\left(1-\frac{c}{S^2}\right)X_i$$ has uniformly smaller risk for all $\theta_i$, where $S^2=\sum_{i=1}^kX_i^2$, the constant $0<c<2(k-2)$, and the risk is defined as
$$R(\bm\theta,\widehat{\bm\theta})=E\left
[\sum_{i=1}^k(\widehat\theta_i-\theta_i)^2\right]$$ with $\bm\theta=(\theta_1,\cdots,\theta_k)^T$ and $\widehat{\bm\theta}=(\widehat\theta_1,\cdots,\widehat\theta_k)^T.$ For a simple proof and explanation see Stigler (1990).
This shows that although each parameter $\theta_i$ is unrelated to $(\theta_j,X_j)$ for $j\neq i$, by the information of all $X_1,\cdots,X_k$, instead of single $X_i$, James-Stein estimator is better than the ``ordinary" estimator $\widehat\theta_i^0=X_i$, which only uses the information of $X_i$.

It can be seen that the situation of our transfer learning is similar to the above example.
The main difference between our transfer learning models and the above example is that in our transfer learning GLMs, the relationship between parameters can be established artificially by historical data or historical characteristics of permanent variables. Note that the historical data in the two models are independent of each other. Thus, in our framework, the two models are unrelated to each other, and the corresponding historical data of the two models are independent of each other as well, implying that the model conditions are basically in accord with those in Stein's paradox. The main differences from the James-Stein estimation method are that our transfer learning estimator is not shrunk (see the proof of Theorem 3.1), and the information of source model $P$ is transferred into the target model $Q$ to enhance the convergence rate of parameter estimation, instead of reducing the risk of parameter estimation.

\subsection{The parameter relationship in general non-linear GLMs}

We consider general non-linear GLMs in which, unlike the condition in Theorem 2.3, the covariates $X^P$ may not be normally distributed.

\subsubsection{Iterative algorithm}

In the general non-linear GLMs, the relationship between the parameters given in Lemma 2.1 has no explicit representation if there is no additional conditions. For the simplicity of notation, we only consider the case where all variables are real-valued.
Note that the empirical version of the result in Lemma 2.1 can be expressed as
\begin{eqnarray*}\frac{1}{n_P}\sum_{i=1}^{n_P}X^P_i \dot{G}_P(\bm\beta X^P_i)=\Lambda \frac{1}{n_Q}\sum_{i=1}^{n_Q}  X^Q_i \dot{G}_Q(\bm\gamma X^Q_i+
\bm\theta Z^Q_i).\end{eqnarray*}
Denote by $\widehat{\bm\beta}=\widehat {\bm s}(\bm\gamma,\bm\theta)$ the solution of $\bm\beta$ via the above equation, namely, the numerical solution of the implicit function $\bm\beta=s(\bm\gamma,\bm\theta)$. We already have some methods for approximately representing the above implicit function. For example, by the Wu-Ritt method, an approximate representation is
$$\widehat {\bm s}(\bm\gamma,\bm\theta)\approx {\bm s}_m={\bm s}_m(\bm\gamma,\bm\theta)={\bm s}_0+\psi(\bm\gamma,\bm\theta,{\bm s}_{m-1}).$$ where $m$ is a relatively large integer, $s_0$ is an initial estimator of $\bm\beta$, for example the MLE obtained by the data from model $P$, and
$$\psi(\bm\gamma,\bm\theta,\bm s)={\bm s}-{\bm s}_0+\frac{\frac{1}{n_P}\sum_{i=1}^{n_P}X^P_i \dot{G}_P({\bm s} X^P_i)-\Lambda \frac{1}{n_Q}\sum_{i=1}^{n_Q}  X^Q_i \dot{G}_Q(\bm\gamma X^Q_i+
\bm\theta Z^Q_i)}{\frac{1}{n_P}\sum_{i=1}^{n_P}(X^P_i)^2 \ddot{G}_P({\bm s} X^P_i)}.$$ When the iteration times $m$ is large enough, the above approximation has satisfactory accuracy.

\subsubsection{Explicit representation}

For explicit representation, we need the additional condition that the historical data or historical characteristics of $Z^Q$ are available as well.
Let $f_P(X^P)$ and $f_Q(X^Q,Z^Q)$ be the marginal density and joint density functions of $X^P$ and $(X^Q,Z^Q)$ respectively, $\dot{f}_P(X^Q)$ be the derivative of $f_P(X^P)$, and $\dot{f}_{Q,X^Q}(X^Q,Z^Q)$ be the derivative of $f_Q(X^Q,Z^Q)$ with respect to $X^Q$. Similar to the results in Lemma 2.1, by the properties of expectation and variance of exponential family distribution, we get the following representations.

\noindent{\bf Lemma A.1.} {\it Under non-linear GLMs, if $$f^2_P(X^P)E[Y^P|X^P]\rightarrow 0 \ \mbox{ and } \ f^2_Q (X^Q,Z^Q)E[Y^Q|X^Q,Z^Q]\rightarrow 0$$ as $\|X^P\|_2\rightarrow\infty$ and $\|(X^Q,Z^Q)\|_2\rightarrow\infty$, then
\begin{eqnarray*}&&-2E[Y^P\dot f_P(X^P)]=\bm\beta E[f_P(X^P) \ddot{G}_P(\bm\beta^T X^P)],\\&&-2E[Y^Q\dot f_{Q,X^Q}(X^Q,Z^Q)]=\bm\gamma E[f_Q(X^Q,Z^Q) \ddot{G}_P(\bm\gamma^T X^Q+\bm\theta^T Z^Q)].\end{eqnarray*}\vspace{-3ex}}

The conditions in the lemma are of tail constraints on the density functions and regression functions, which are mild and are commonly used in the existing literature (see, e.g., Powell et al., 1989; H\"rdle et al., 2004). The lemma indicates that the linear correlation $E[\dot f_P(X^P)Y^P]$ between $\dot f_P(X^P)$ and $Y^P$, and the linear correlation $E[\dot f_{Q,X^Q}(X^Q,Z^Q)Y^Q]$ between $\dot f_{Q,X^Q}(X^Q,Z^Q)$ and $Y^Q$ can capture the information of regression coefficients $\bm\beta$ and $\bm \gamma$, respectively. We then use them to establish the relationship between the regression coefficients. Let $\dot f^j_P(X^P)$ and $\dot f^j_{Q,X^Q}(X^Q,Z^Q)$ be the $j$-th components of $\dot f_P(X^P)$ and $\dot f_{Q,X^Q}(X^Q,Z^Q)$ respectively, and write
$$\Omega=\mbox{diag}\left(\frac{E\left[Y^P\dot f^1_P(X^P)\right]}{E\left[Y^Q\dot f^1_{Q,X^Q}(X^Q,Z^Q)\right]},\cdots,\frac{E\left[Y^P\dot f^{d_1}_P(X^P)\right]}{E\left[Y^Q\dot f^{d_1}_{Q,X^Q}(X^Q,Z^Q)\right]}\right).$$
It will be seen later that the matrix $\Omega$ can be estimated by historical data or historical characteristics. Then we suppose it is known in advance or has been estimated by historical data or historical characteristics. Unlike $\Lambda$, here the estimation of $\Omega$ needs the density functions of $X^P$ and $(X^Q,Z^Q)$. A motivating example is that, according to the privacy policy, the raw data of $Z^Q$ is confidential before a time point. But their some characteristics, such as expectations and densities, are not confidential. Thus, we can use these historical characteristics to determine $\Omega$.

We have the following theorem.

\noindent{\bf Theorem A.2.} {\it If both $P$ and $Q$ are non-linear GLMs, the tail conditions given in Lemma A.1 hold, $E\left[Y^Q\dot f^j_{Q,X^Q}(X^Q,Z^Q)\right]\neq 0$ for $j=1,\cdots,d_1$, $E[f_P(X^P) \ddot{G}_P(\bm\beta^T X^P)]\neq 0$, and $\Omega$ is known in advance or has been estimated by historical data, then, the model parameters satisfy
\begin{eqnarray*}\bm\beta=\frac{E[f_Q(X^Q,Z^Q) \ddot{G}_Q(\bm\gamma^T X^Q+\bm\theta^T Z^Q)]}{E\{f_P(X^P)Var[Y^P|X^P]\}}
\Omega\, \bm\gamma\,.\end{eqnarray*}\vspace{-3ex}}

By the combination between the parameters, we can get the transfer learning likelihood estimator as in (2.14), and the main theoretical property: the transfer learning estimator $\widehat{\bm\gamma}$ is $\sqrt n$-consistent if $b_j^2>c_j(\rho)$ for some $c_j(\rho)>0$, $j=1,\cdots,d_1$, where $b_j=\frac{E\left[Y^P\dot f^j_P(X^P)\right]}{E\left[Y^Q\dot f^j_{Q,X^Q}(X^Q,Z^Q)\right]}$.

\subsection{The estimators of correlation-ratio matrices}

\subsubsection{The estimation method for $\Lambda$ by historical data} As an example, the estimator $\widehat a$ of $a=\frac{E[X^PY^P]}{E[X^QY^Q]}$ is considered here; the estimators for the others are similar. In this case, all the relevant variables are real-valued. We suppose that $(\widetilde X^P_i,\widetilde Y^P_i),i=1,\cdots,\widetilde n_{P}$, and  $(\widetilde X^Q_i,\widetilde Y^Q_i),i=1,\cdots,\widetilde n_{Q}$, are historical data respectively of $(X^P,Y^P)$ and $(X^Q,Y^Q)$.
It is known that $\frac{\frac{1}{\widetilde n_{P}}\sum_{i=1}^{\widetilde n_P}\widetilde Y^P_i\widetilde X^P_i}
{\frac{1}{\widetilde n_{Q}}\sum_{i=1}^{\widetilde n_Q}\widetilde Y^Q_i\widetilde X^Q_i}$ is a biased estimator of the ratio $\frac{E[Y^PX^P]}{E[Y^QX^Q]}$ (Mullen, 1967). According to the bias-correction estimation for the ratio of parameters (Uno, 2003), we consider the following bias-corrected estimator:
$$\widehat a(c)=\frac{\frac{1}{\widetilde n_{P}}\sum_{i=1}^{\widetilde n_P}\widetilde Y^P_i\widetilde X^P_i}
{\frac{1}{\widetilde n_{Q}}\sum_{i=1}^{\widetilde n_Q}\widetilde Y^Q_i\widetilde X^Q_i}
-c,$$ where $c\geq 0$ is a bias-correction parameter of order from $O(1/\widetilde n_P)$ to $O(1/\widetilde n_Q)$.
For a given $c$, we get the leave-one-out transfer estimators of $\bm\gamma$, $\bm\theta$ and $\bm s(\bm\gamma,\bm\theta)$ respectively as $\widehat{\bm\gamma}_{-i}(c)$, $\widehat{\bm\theta}_{-i}(c)$ and $\widehat{\bm s}(\widehat{\bm\gamma}_{-i}(c),\widehat{\bm\theta}_{-i}(c))$ by the method proposed. According to C-V rule, we choose $c$ by minimizing
\begin{eqnarray*}CV(c)&=&\widehat w_P\sum_{i=1}^{\widetilde n_P}\left ( \widetilde Y_i^P-g^{-1}\left(\widehat{\bm s}^T(\widehat{\bm\gamma}_{-i}(c),\widehat{\bm\theta}_{-i}(c))\widetilde X^P_i\right)
\right)^2\\&&+\widehat w_Q\sum_{i=1}^{\widehat n_Q}\left ( \widetilde Y_i^Q-g^{-1}\left(\widehat{\bm\gamma}_{-i}^T(c)\widetilde X^Q_i+\widehat w_Q
\widehat{\bm\gamma}^T_{-i}(c) \widetilde Z^Q_i\right)\right)^2.\end{eqnarray*} Denote by $\widehat c$ the solution to the above optimization problem. Then, the final estimator of $a$ is $\widehat a(\widehat c)$.

\subsubsection{The estimation method for $\Omega$ by historical data}
The numerator $E\left[Y^P\dot f^j_P(X^P)\right]$ of each diagonal element of $\Omega$ can be accurately estimated because the sample size of $P$ is large. In the following, we focus mainly on the estimation method for the denominator $E\left[Y^Q\dot f^j_{Q,X^Q}(X^Q,Z^Q)\right]$, equivalently for the nonparametric estimator of $\dot f^j_{Q,X^Q}(X^Q,Z^Q)$. To achieve the goal, in addition to historical data $(\widetilde X^Q_i,\widetilde Y^Q_i),i=1,\cdots,\widetilde n_{Q}$, some additional historical data $(\widetilde X^Q_i,\widetilde Z^Q_i), i=\widetilde n_{Q}+1,\cdots,\widetilde n_{Q}+ n^*_{Q}$, are required as well. Here the sample size $ n_{Q}^*$ is relatively small because it is difficult to obtain the observation values of $Z^Q$.

For simplicity of notation, here we only consider the case where all the variables are real-valued. To sufficiently use the historical data, we rewrite the join density $f_{Q}(X^Q,Z^Q)$ as $f_{Q}(X^Q,Z^Q)=f_{Q}(X^Q)f_Q(Z^Q|X^Q)$. Note that $f_{Q}(X^Q)$ can be accurately estimated by all the data $\widetilde X^Q_i, i=1,\cdots,\widetilde n_{Q}+ n^*_{Q}$; the estimator is denoted by $\widehat f_{Q}(X^Q)$. The remaining task is  to construct the estimator of the conditional density $f_Q(Z^Q|X^Q)$. We suggest the two-step estimation method (Hansen, 2004) for realizing this goal.
Let
$m(X^Q)=E[\widetilde Z^Q_i|\widetilde X^Q_i=X^Q]$ so that $\widetilde Z^Q_i=m(\widetilde X^Q_i)+e_i$ and $e_i$ is a regression error. Denote by $\psi(e|X^Q)$ the conditional density of $e_i$ given $\widetilde X^Q_i=X^Q$. We have
$$f_Q(Z^Q|X^Q)=\psi(Z^Q-m(X^Q)|X^Q).$$ Here $m(X^Q)$ can be estimated by for example the N-W estimation method as
$$\widehat m(X^Q)=\frac{\sum_{i=\widetilde n_Q+1}^{\widetilde n_Q+n_Q^*}Z^Q_iK_{h_0}\{(X^Q_i-X^Q)/h_0\}}{\sum_{i=\widetilde n_Q+1}^{\widetilde n_Q+n_Q^*}K_{h_0}\{(X^Q_i-X^Q)/h_0\}},$$ where $K(\cdot)$ is a kernel function and $h_0$ is the bandwidth. With the estimator, the residuals are $\widehat e_i=Z^Q_i-\widehat m(X^Q_i)$. Then, the two-step estimator of $\psi(\cdot)$ is
$$\psi(e|X^Q)=\frac{\sum_{i=\widetilde n_Q+1}^{\widetilde n_Q+n_Q^*}K_{h_1}\{(X^Q_i-X^Q)/h_1\}K_{h_2}\{(\widehat e_i-e)/h_2\}}{\sum_{i=\widetilde n_Q+1}^{\widetilde n_Q+n_Q^*}K_{h_1}\{(X^Q_i-X^Q)/h_1\}},$$ and consequently, the two-step estimator of $f_Q(Z^Q|X^Q)$ is
$$\widehat f_Q(Z^Q|X^Q)=\frac{\sum_{i=\widetilde n_Q+1}^{\widetilde n_Q+n_Q^*}K_{h_1}\{(X^Q_i-X^Q)/h_1\}K_{h_2}\{(\widehat e_i-(Z^Q-\widehat m(X^Q)))/h_2\}}{\sum_{i=\widetilde n_Q+1}^{\widetilde n_Q+n_Q^*}K_{h_1}\{(X^Q_i-X^Q)/h_1\}}.$$
By the above estimators, we can get the estimator of the derivative of
$\widehat f_{Q}(X^Q)\widehat f_Q(Z^Q|X^Q)$ as $\dot{\widehat f_{Q}}(X^Q)\widehat f_Q(Z^Q|X^Q)+\widehat f_{Q}(X^Q)\dot{\widehat f_Q}(Z^Q|X^Q)$, where $\dot{\widehat f_{Q}}(X^Q)$ and $\dot{\widehat f_Q}(Z^Q|X^Q)$ are respectively the derivatives of $\widehat f_{Q}(X^Q)$ and $\widehat f_Q(Z^Q|X^Q)$ with respect to $X^Q$.

After the estimators of the numerator and denominator are obtained, the estimator of $\Omega$ can be constructed by the bias-correction estimation for the ratio proposed in the previous subsection.

\subsection{Accurate transfer learning for non-linear models}

We consider the scenario where the source model $P$ is still a GLM as (2.1), but the target model $Q$ is the following general non-linear model:
\begin{eqnarray}\label{(Q-nonlinear)}E[Y^Q|X^Q,Z^Q]=g_Q^{-1}(X^Q,Z^Q,\bm\alpha),\end{eqnarray} where $\bm\alpha=(\alpha^1,\cdots,\alpha^{d_1})^T$ is a parametric vector with dimension $d_1$, and the pre-specified function $g_Q(\cdot,\cdot,\bm\alpha)$ is non-linear in $\bm\alpha$. Under an exponential family distribution, the log-likelihood in model $Q$ has the form: $$l_Q(\bm\gamma,\bm\theta|Y^Q,X^Q,Z^Q)\propto Y^Qg_Q(X^Q,Z^Q,\bm\alpha) -G_Q(g_Q(X^Q,Z^Q,\bm\alpha)).$$ The true value of $\bm\alpha$ is defined as the solution to the following likelihood equation:
\begin{eqnarray}\label{(Q-nonlinear equation)} E[Y^Q\dot{g}_{Q,\bm\alpha}(X^Q,Z^Q,\bm\alpha)-  \dot{G}_Q({g}_Q(X^Q,Z^Q,\bm\alpha))\dot{g}_{Q,\bm\alpha}(X^Q,Z^Q,\bm\alpha)]=0,\end{eqnarray} where $\dot{g}_{Q,\bm\alpha}(X^Q,Z^Q,\bm\alpha)$
is the derivative of $g_Q(X^Q,Z^Q,\bm\alpha)$ with respect to $\bm\alpha$.
Together with likelihood equation (2.2), we have the following accurate representation:
\begin{eqnarray}\label{(Q-nonlinear equation-1)}E[X^P \dot{G}_P(\bm\beta X^P)]=\Xi(\bm\alpha)  E[\dot{g}_{Q,\bm\alpha}(X^Q,Z^Q,\bm\alpha) \dot{G}_Q(g_Q(X^Q,Z^Q,\bm\alpha)].\end{eqnarray} Here the correlation-ratio matrix $$\Xi(\bm\alpha) =\mbox{diag}\left(\frac{E[Y^PX^{1P}]}{E[Y^Q\dot{g}_{Q,\bm\alpha^1}
(X^Q,Z^Q,\bm\alpha)]}\cdots,\frac{E[Y^PX^{d_1P}]}
{E[Y^Q\dot{g}_{Q,\alpha^{d_1}}
(X^Q,Z^Q,\alpha)]}\right),$$ which is known in advance or has been consistently estimated by historical data. The above builds an accurate relationship between the parameters in the two models.
The main difference from Lemma 2.1 is that here the correlation-ratio $\Xi(\bm\alpha)$ depends on unknown parameter $\bm\alpha$. By differential mean value theorem, we can see that here $\Xi(\bm\alpha)$ describes an approximate correlation-ratio of the linear correlations between response $Y^P$ and covariate $X^P$, and response $Y^Q$ and ``covariate" $\dot{g}_{Q,\bm\alpha}(X^Q,Z^Q,\bm\alpha^0)$ respectively in the two models.

Consider the special regime where model $P$ is a linear model but model $Q$ is a general non-linear model as in (\ref{(Q-nonlinear)}). By representation (\ref{(Q-nonlinear equation-1)}), we get an explicit expression for $\bm\beta$ as
\begin{eqnarray}\label{(nonlinear representation)}\bm\beta=(E[X^P(X^P)^T])^{-1}\Xi(\bm\alpha) E[\dot{g}_{Q,\bm\alpha}(X^Q,Z^Q,\bm\alpha) \dot{G}_Q(g_Q(X^Q,Z^Q,\bm\alpha)].\end{eqnarray} Unlike the representation in Theorem 2.2, in the above, $\bm\beta$ is a non-linear function of $\bm\alpha$. Moreover, similar to Theorem 2.3, suppose that model $P$ is a general GLM as in (2.1) but $X^P\sim N(0,I_{d_1})$, we have the closed expression for $\bm\beta$ as
\begin{eqnarray}\label{(nonlinear representation-normal)}\bm\beta=(E\{Var[Y^P|X^P])^{-1}\Xi(\bm\alpha) E[\dot{g}_{Q,\bm\alpha}(X^Q,Z^Q,\bm\alpha) \dot{G}_Q(g_Q(X^Q,Z^Q,\bm\alpha)].\end{eqnarray}
In the representations, (\ref{(nonlinear representation)}) and  (\ref{(nonlinear representation-normal)}), we use the same correlation-ratio matrix $\Xi(\bm\alpha)$ to combine the parameters in the two models.

We further consider general case where model $P$ is a general GLM as in (2.1) and model $Q$ is a general non-linear model as in (\ref{(Q-nonlinear)}), without the normality condition of covariate $X^P$. By the same argument as used in the proofs of Lemma A.1 and Theorem A.2, we get the following representation:
\begin{eqnarray}\label{(nonlinear representation-1)}\nonumber\bm\beta&=&\left(f_P(X^P)Var[Y^P|X^P]\}\right)^{-1}
\Omega\\&&\times E[f_Q(X^Q,Z^Q) \ddot{G}_Q(X^Q, Z^Q,\bm\alpha)\dot{g}_{Q,\bm\alpha}(X^Q,Z^Q,\bm\alpha)],\end{eqnarray} where $\Omega$ is defined as the same as in the above subsection.

The closed representations in (\ref{(nonlinear representation)}), (\ref{(nonlinear representation-normal)}) and (\ref{(nonlinear representation-1)}) and their estimated versions can be unified expressed respectively as
\begin{eqnarray*}\bm\beta=\bm s(\bm\alpha) \ \mbox{ and } \ \bm\beta=\widehat{\bm s}(\bm\alpha)\ \mbox{ uniformly for } \bm\alpha\in \cal A,\end{eqnarray*} where $\cal A$ is the parameter space of $\bm\alpha$.
Consequently, the estimated cr-TLL can be expressed as
\begin{eqnarray}\label{(nonlinear likelihood)}\nonumber\hspace{-2ex}\widehat l_{(P,Q)}(\bm\alpha)&=&\widehat w_P\sum_{i=1}^{n_P}\left(Y^P_iX^P_i {\widehat {\bm s}}(\bm\alpha) -X^P_iG(\widehat {\bm s}(\bm\alpha) )\right)\\&& +\widehat w_Q\sum_{i=1}^{n_Q}\left(Y^Q_i{g}_{Q}(X^Q_i,Z^Q_i,\bm\alpha)-  {G}_Q({g}_Q(X^Q_i,Z^Q_i,\bm\alpha))\right)=0.\end{eqnarray}
The transfer learning likelihood estimator $\widehat{\bm\alpha}$ is then obtained by maximizing the above likelihood, namely,
\begin{eqnarray}\label{(nonlinear-estimator)}
\widehat{\bm\alpha}
= \arg\max_{\bm\alpha\in \cal A} \widehat l_{(P,Q)}(\bm\alpha).\end{eqnarray}

When the function $g(\cdot,\cdot,\cdot)$ and above likelihood function satisfy the regularity conditions given in  Wei (1998), the theoretical properties of $\widehat{\bm\alpha}$ are similar to those in the previous section, that is to say, with a significant correlation-ratio, the estimator can achieve the global convergence rate of order $O_p(1/\sqrt{n})$; the details are omitted here.

\subsection{Regularity conditions}

Theorem 3.3 needs the following regularity conditions:

\begin{itemize}\item [C6.] Parameter spaces $\Gamma$ and $\Theta$ of $\bm\gamma$ and $\bm\theta$ are bounded, convex and closed subsets of $\mathbb{R}^{d_1}$ and $\mathbb{R}^{d_2}$, respectively.
\item [C7.] The functions $g_P(\cdot)$, $g_Q(\cdot)$ and $\bm s(\cdot)$ are twice continuously differentiable, and functions $G_P(\cdot)$ and $G_Q(\cdot)$ are differentiable up to the fourth order.
\item [C8.] The matrix $$w_P\dot{\bm s}^T(\bm\alpha^0)\sum_{i=1}^{n_P}\sum_{j=1}^{d_1}\sigma_{P, i}^2 X_i^{P}(X_i^{P})^T\dot{\bm s}(\bm\alpha^0)+w_Q\sum_{i=1}^{n_Q}\sigma_{Q, i}^2((X_i^Q)^T,(Z_i^Q)^T)^T((X_i^Q)^T,(Z_i^Q)^T)$$ has full rank when $n_P$ and $n_Q$ are large enough.
\item [C9.] The minimum eigenvalue of $F_{n}(\bm\alpha^0)$ satisfies $\lambda_{\min}F_{n}(\bm\alpha^0)\rightarrow\infty$.
\item [C10.] $\max_{\bm\alpha\in N_n(\delta)}\|(F_{n}^T(\bm\alpha^0))^{-1/2}F_{n}(\bm\alpha)
    (F_{n}(\bm\alpha^0))^{-1/2}-I\|\rightarrow 0$, where $N_n(\delta)=\{\bm\alpha:\|F_n^{1/2}(\bm\alpha-\bm\alpha^0)\|\leq \delta\}$.
    \end{itemize}
These are commonly used conditions for the asymptotic properties of the maximum likelihood estimation in GLMs; for the rationality and the roles of the conditions in the proof of the following theorem, see for example  Fahrmeir and Kaufmann (1985). We need the boundedness of $\Gamma$ and $\Theta$ only for the simplicity of the proof of Theorem 3.3.

\subsection{Proofs of Lemmas and Theorems}

\noindent {\it Proof of Lemma 2.1.} By the likelihood equations (2.2) and (2.4), we have
\begin{eqnarray*}E[Y^PX^{jP}]=E[X^{jP} \dot{G}_P(\bm\beta^T X^P)], \ E[Y^QX^{jQ}]=E[X^{j Q}\dot{G}_Q(\bm\gamma^TX^Q+
\bm\theta^T Z^Q)]\end{eqnarray*} for $j=1,\cdots,d_1$. This implies the result of the lemma. $\square$

\noindent {\it Proof of Theorem 2.2.} Note that for linear regression, $G_P(u)$ and $G_Q(u)$ are proportional to $u^2/2$. Then, Lemma 2.1 leads to the result of the theorem. $\square$

\noindent {\it Proof of Theorem 2.3.} Because $X^P$ is normally distributed, by the First-Order Stein's Identity (Stein, 1986; Yang et al., 2017), we have
$$E[X^P\dot G_P(\bm\beta^TX^P)]=E[\ddot G_P(\bm\beta^TX^P)]\bm\beta=E\{Var[Y^P|X^P]\}\bm\beta.$$ By the result and Lemma 2.1, we get the first result of the theorem. Similarly, if $X^Q$ is also normally distributed, we have $$E[X^Q\dot G_Q(\bm\gamma^TX^Q+\bm\theta^TZ^Q)]=E[\ddot G_Q(\bm\gamma^TX^Q+\bm\theta^TZ^Q)]\bm\gamma.$$ By the above results and Lemma 2.1, we get the second result of the theorem. $\square$

\noindent {\it Proof of Lemma A.1.} By the expectation property of exponential family distribution, we have
$E[Y^P|X^P]=\dot{G}_P(\bm\beta^T X^P),$ and then
$$\dot E[Y^P|X^P]=\bm\beta\ddot{G}_P(\bm\beta^T X^P),$$ where $\dot E[Y^P|X^P]$ is the derivative of $E[Y^P|X^P]$ with respect to $X^P$. Consequently, $$E\{f_P(X^P)\dot E[Y^P|X^P]\}=\bm\beta E[f_P(X^P) \ddot{G}_P(\bm\beta^T X^P)].$$ Note that partial integration yields
$$E\{f_P(X^P)\dot E[Y^P|X^P]\}=-2E\{\dot f_P(X^P)E[Y^P|X^P]\}=-2E[\dot f_P(X^P)Y^P]$$ if $f^2_P(X^P)E[Y^P|X^P]\rightarrow 0$ as $\|X^P\|_2\rightarrow\infty$. We then get the first representation in the lemma. The proof for the second result is similar.   $\square$

\noindent {\it Proof of Theorem A.2.} By Lemma A.1, and the property of variance, $\ddot G_P(\bm\beta^T X^P)=Var[Y^P|X^P]$, we can prove the lemma.
$\square$

\noindent{\it Proof of Theorem 3.1.} Denote \begin{eqnarray*}&&{\cal A}_n=\frac{1}{n}\left(w_P\sum_{i=1}^{n_P}(X_i^P)^2\left(\begin{array}{ll}  a^2& ab\\ab &b^2\end{array}\right)+w_Q\sum_{i=1}^{n_Q}\left(\begin{array}{ll} (X^Q_i)^2&X_i^QZ_i^Q\\ X^Q_iZ^Q_i&(Z_i^Q)^2\end{array}\right)\right), \\
&&{\cal B}_n=\frac{1}{n}\left(
w_P\sum_{i=1}^{n_P}Y^P_iX^P_i
\left(\begin{array}{cc}a\\ b\end{array}\right) +w_Q\sum_{i=1}^{n_Q}Y_i^Q\left(\begin{array}{ll} X^Q_i\\ Z^Q_i\end{array}\right)\right).\end{eqnarray*} Note that $E[Y_i^P|X^P_i]=\bm\beta X_i^P$, $E[Y_i^Q|X^Q_i,Z_i^Q]=\bm\gamma X_i^Q+\bm\theta Z_i^Q$, $\bm\beta=a\bm\gamma+b\bm\theta$, and the estimation consistency: $\widehat a=a+o_p(1), \widehat b=b+o_p(1), \widehat w_P=w_P+o_p(1)$ and $\widehat w_Q=w_Q+o_p(1)$. By these representations, together with Slutsky lemma, it can be verified that the estimators (2.13) can be expressed as
\begin{eqnarray*}
\sqrt n\left(\begin{array}{ll}\widehat{\bm\gamma}
\\\widehat{\bm\theta}\end{array}
\right)=\sqrt n\left(\begin{array}{ll}\widetilde{\bm\gamma}
\\ \widetilde{\bm\theta}\end{array}
\right)+\sqrt n\,o_p({\cal A}^{-1}_n){\cal B}_n+\sqrt n\,{\cal A}^{-1}_no_p({\cal B}_n)+\sqrt n\,o_p({\cal A}^{-1}_n)o_p({\cal B}_n),\end{eqnarray*} where $\sqrt n\left(\begin{array}{ll}\widetilde{\bm\gamma}
\\ \widetilde{\bm\theta}\end{array}
\right)=\sqrt n\,{\cal A}^{-1}_n{\cal B}_n$, and the last three terms on the right-hand side are of order $o_p(1)$.
Then, asymptotically, $\sqrt n\left(\begin{array}{ll}\widehat{\bm\gamma}
\\\widehat{\bm\theta}\end{array}
\right)$
is identically distributed as $\sqrt n\left(\begin{array}{ll}\widetilde{\bm\gamma}
\\ \widetilde{\bm\theta}\end{array}
\right)$. Because ${\cal A}^{-1}_n$ tends to a positive definite matrix and ${\cal B}_n$ is an average of independent random variables, $\sqrt n\,{\cal A}^{-1}_n{\cal B}_n$ is normally distributed, its expectation is
$$E\left[\sqrt n\,{\cal A}^{-1}_n{\cal B}_n\right]=\sqrt n\left(\begin{array}{ll}{\bm\gamma}
\\ {\bm\theta}\end{array}
\right),$$
and, asymptotically, the covariance satisfies
\begin{eqnarray*} Cov\left[\sqrt n\,{\cal A}^{-1}_n{\cal B}_n\right]&\approx&\left(n_Pw_P\left(\begin{array}{cc}a^2& ab\\ ab& b^2\end{array}\right)+n_Qw_Q\left(\begin{array}{cc} 1&\rho_{X^QZ^Q}\\ \rho_{X^QZ^Q}&1\end{array}\right)\right)^{-1}.\end{eqnarray*} Note that it can be directly verified that $n_P a^2+n_Q-n_P  a^2\frac{\rho^2_{X^QZ^Q}\left(1+\frac{\tau}{a^2}\right)^2}
{\rho^2_{X^QZ^Q}+\frac{\tau}{a^2}}>0$. Then, by the condition of $w_p=w_Q=1/2$ given in the theorem, the relationship $b=a\rho_{X^QZ^Q}$ and the above representation, we have that, asymptotically,
\begin{eqnarray*}Var[\sqrt n\,\widetilde{\bm\gamma}]&\approx &n\left(n_P a^2+n_Q-\frac{\left(n_P a b+n_Q\rho_{X^QZ^Q}\right)^2}{n_P b^2+n_Q}\right)^{-1}\\&=&n\left(n_P a^2+n_Q-n_P  a^2\frac{\rho^2_{X^QZ^Q}\left(1+\frac{n_Q}{n_Pa^2}\right)^2}
{\rho^2_{X^QZ^Q}+\frac{n_Q}{n_Pa^2}}\right)^{-1}
\\&\approx &n\left(n_P a^2+n_Q-n_P  a^2\frac{\rho^2_{X^QZ^Q}\left(1+\frac{\tau}{a^2}\right)^2}
{\rho^2_{X^QZ^Q}+\frac{\tau}{a^2}}\right)^{-1}.
\end{eqnarray*}
By combining the results above, we can prove the first result of the theorem.

If $|\rho_{X^QZ^Q}|<1/\sqrt 2$ and $a^2>\tau\rho^2_{X^QZ^Q}/(1-2\rho^2_{X^QZ^Q})$, we can verify
\begin{eqnarray*}\rho^2_{X^QZ^Q}/(1-2\rho^2_{X^QZ^Q})>0 \ \mbox{ and }\ \frac{\rho^2_{X^QZ^Q}\left(1+\frac{\tau}{a^2}\right)^2}
{\rho^2_{X^QZ^Q}+\frac{\tau}{a^2}}<1.
\end{eqnarray*}
Consequently,
\begin{eqnarray*}Var[\sqrt n\,\widetilde{\bm\gamma}]\approx n\left(n_P a^2\left(1-\frac{\rho^2_{X^QZ^Q}\left(1+\frac{\tau}{a^2}\right)^2}
{\rho^2_{X^QZ^Q}+\frac{\tau}{a^2}}\right)+n_Q\right)^{-1}
=O\left(1\right).
\end{eqnarray*}
Then, the proofs for the second and third conclusions are completed.
$\square$

\noindent{\it Proof of Proposition 3.2.} By the same argument as used in the proof of Theorem 3.1, we only need to calculate the variance of the estimator $\widehat{\bm\gamma}$. Note that $b=a\rho_{X^QZ^Q}$. Then, it can be seen that asymptotically
\begin{eqnarray*}Var[\widehat{\bm\gamma}]&\approx & \left(n_P b^2+n_Q-\frac{\left(n_P a b+n_Q\rho_{X^QZ^Q}\right)^2}{n_P a^2+n_Q}\right)^{-1}\\&\approx&\left(n_P b^2+n_Q-n_P  b^2\frac{\left(1+\frac{\tau}{a^2}\right)^2}
{1+\frac{\tau}{a^2}}\right)^{-1}\\
&=&\frac{1}{n_Q(1-\rho_{X^QZ^Q})}.
\end{eqnarray*} The proof is completed. $\square$

\noindent{\it Proof of Theorem 3.3.} It follows from the consistency given in (\ref{(representation-1_1)}) that
\begin{eqnarray*}\frac{1}{n}\widehat l_{(P,Q)}(\bm\alpha)=\frac{1}{n}l_{(P,Q)}(\bm\alpha)+
\frac{1}{n}\sum_{i=1}^{n_P}\left(Y^P_i {\bm s}^T(\bm\alpha)X^P_i -{G}_P({\bm s}^T(\bm\alpha) X^P_i)\right)O_p\left(\epsilon_n\right),\end{eqnarray*} where $\epsilon_n=1/\sqrt{n_m}$. In the above, the $\frac{1}{n}\sum_{i=1}^{n_P}\left(Y^P_i{\bm s}^T(\bm\alpha) X^P_i -{G}_P({\bm s}^T(\bm\alpha) X^P_i)\right)O_p\left(\epsilon_n\right)$ is of order $O_p\left(\frac{\sqrt{n_P}\,\epsilon_n}{n}\right)$. This result and the regularity conditions lead to that the objective function $\frac{1}{n}\widehat l_{(P,Q)}(\bm\alpha)$ satisfies the conditions of Theorem 5.7 of  Van der Vaart (2000), ensuring that the estimator $\widehat{\bm\alpha}$ is consistent in probability. We then have
$$0=\frac{1}{n}\dot{\widehat l}_{(P,Q)}(\bm\alpha^0)+\frac{1}{n}\ddot{\widehat l}_{(P,Q)}(\bm\alpha^0)(\widehat{\bm\alpha}-\bm\alpha^0)^T+
O_p(\|\widehat{\bm\alpha}-\bm\alpha^0\|^2).$$
Thus, asymptotically, $\sqrt n(\widehat{\bm\alpha}-\bm\alpha^0)$ is identically distributed as $$-\left(\frac{1}{n}\ddot{\widehat l}_{(P,Q)}(\bm\alpha^0)\right)^{-1}\frac{1}{\sqrt n}\dot{\widehat l}_{(P,Q)}(\bm\alpha^0).$$ Asymptotically, the above is identically distributed as $$-\left(\frac{1}{n}\ddot{ l}_{(P,Q)}(\bm\alpha^0)\right)^{-1}\frac{1}{\sqrt n}\dot{ l}_{(P,Q)}(\bm\alpha^0).$$
By the above results and Theorem 3 of  Fahrmeir and Kaufmann (1985), the transfer learning estimator $\widehat{\bm\alpha}$ is normally distributed asymptotically as that given in Theorem 3.3. $\square$

\noindent{\it Proof of Corollary 3.4.} It follows from Theorem 3.4 that the asymptotic variance of $\widehat{\bm\gamma}_1$ can be expressed as
\begin{eqnarray*} v_{{\bm\gamma}_1}=\left(n_P\dot{\bm s}^T_1\Sigma^P\dot{\bm s}_1+n_Q\sigma^Q_{11}-n_Pr\right)^{-1}.\end{eqnarray*} We then get the first result of the corollary. For the second the result,
by the result above, we only need to prove \begin{eqnarray*}\dot{\bm s}^T_1\Sigma^P\dot{\bm s}_1-r>0.\end{eqnarray*}
By the definition of $r$, the conditions in the theorem and the Cauchy-Schwarz inequality, we have
\begin{eqnarray*}&&r=(\bm \dot{\bm s}_1^T\Sigma^P\dot{\bm s}_{(-1)}+\tau\bm (\bm\sigma^Q_{12})^T)(A+\tau \Sigma^Q_{22})^{-1}(\dot{\bm s}_{(-1)}^T\Sigma^P\bm \dot{\bm s}_1+\tau\bm \sigma^Q_{12})\\&&=\bm \dot{\bm s}_1^T\Sigma^P\dot{\bm s}_{(-1)}(A+\tau \Sigma^Q_{22})^{-1}\dot{\bm s}_{(-1)}^T\Sigma^P\bm \dot{\bm s}_1+\tau^2\bm (\bm\sigma^Q_{12})^T(A+\tau \Sigma^Q_{22})^{-1}\bm \sigma^Q_{12}\\&&\ \ \ \ +2\tau\bm \dot{\bm s}_1^T\Sigma^P\dot{\bm s}_{(-1)}(A+\tau \Sigma^Q_{22})^{-1}\bm \sigma^Q_{12}\\&&<\bm \dot{\bm s}_1^T\Sigma^P\dot{\bm s}_{(-1)}(A+\tau \Sigma^Q_{22})^{-1}\dot{\bm s}_{(-1)}^T\Sigma^P\bm \dot{\bm s}_1+\tau^2\epsilon\,\bm \dot{\bm s}_1^T\Sigma^P\dot{\bm s}_{(-1)}(A+\tau \Sigma^Q_{22})^{-1}\dot{\bm s}_{(-1)}^T\Sigma^P\bm \dot{\bm s}_1\\&&\ \ \ \ +2\tau\sqrt \epsilon\,\bm \dot{\bm s}_1^T\Sigma^P\dot{\bm s}_{(-1)}(A+\tau \Sigma^Q_{22})^{-1}\dot{\bm s}_{(-1)}^T\Sigma^P\bm \dot{\bm s}_1\\&&=(1+\tau^2\epsilon+2\tau\sqrt \epsilon)\bm \dot{\bm s}_1^T\Sigma^P\dot{\bm s}_{(-1)}(A+\tau \Sigma^Q_{22})^{-1}\dot{\bm s}_{(-1)}^T\Sigma^P\bm \dot{\bm s}_1\\&&=(1+\tau^2\epsilon+2\tau\sqrt \epsilon)(\bm \dot{\bm s}_1^T\Sigma^P\dot{\bm s}_{(-1)}A^{+}\dot{\bm s}_{(-1)}^T\Sigma^P\bm \dot{\bm s}_1-\bm \dot{\bm s}_1^T\Sigma^P\dot{\bm s}_{(-1)}D\dot{\bm s}_{(-1)}^T\Sigma^P\bm \dot{\bm s}_1)\\&&=\bm \dot{\bm s}_1^T\Sigma^P\dot{\bm s}_{(-1)}A^{+}\dot{\bm s}_{(-1)}^T\Sigma^P\bm \dot{\bm s}_1\\&&\ \ \ \ +\left\{(\tau^2\epsilon+2\tau\sqrt \epsilon)\bm \dot{\bm s}_1^T\Sigma^P\dot{\bm s}_{(-1)}A^{+}\dot{\bm s}_{(-1)}^T\Sigma^P\bm \dot{\bm s}_1-(1+\tau^2\epsilon+2\tau\sqrt \epsilon)\bm \dot{\bm s}_1^T\Sigma^P\dot{\bm s}_{(-1)}D\dot{\bm s}_{(-1)}^T\Sigma^P\bm \dot{\bm s}_1\right\}\\&&\leq\bm \dot{\bm s}_1^T\Sigma^P\dot{\bm s}_{(-1)}A^{+}\dot{\bm s}_{(-1)}^T\Sigma^P\bm \dot{\bm s}_1\\&&\ \ \ \ +\left\{(\tau^2\epsilon+2\tau\sqrt \epsilon)\bm \dot{\bm s}_1^T\Sigma^P\dot{\bm s}_{(-1)}A^{+}\dot{\bm s}_{(-1)}^T\Sigma^P\bm \dot{\bm s}_1-\bm \dot{\bm s}_1^T\Sigma^P\dot{\bm s}_{(-1)}D\dot{\bm s}_{(-1)}^T\Sigma^P\bm \dot{\bm s}_1\right\}\\&&\leq\bm \dot{\bm s}_1^T\Sigma^P\dot{\bm s}_{(-1)}A^{+}\dot{\bm s}_{(-1)}^T\Sigma^P\bm \dot{\bm s}_1,\end{eqnarray*}
where $D=A^{+}-\tau A^{+}((\Sigma^Q_{22})^{-1}+\tau A^{+})^{-1}A^{+}$. We then have
\begin{eqnarray*}&&\dot{\bm s}^T_1\Sigma^P\dot{\bm s}_1-r\\&&>\dot{\bm s}^T_1\Sigma^P\dot{\bm s}_1-\bm \dot{\bm s}_1^T\Sigma^P\dot{\bm s}_{(-1)}A^{+}\dot{\bm s}_{(-1)}^T\Sigma^P\bm \dot{\bm s}_1\\&&=\dot{\bm s}_1^T(\Sigma^P)^{1/2}\left\{I-(\Sigma^P)^{1/2}\dot{\bm s}_{(-1)} A^{+}\dot{\bm s}_{(-1)}^T(\Sigma^P)^{1/2}\right\}(\Sigma^P)^{1/2}\bm \dot{\bm s}_1\\&&\geq0\end{eqnarray*}
because $\dot{\bm s}_1\neq \bm 0$, and the matrix $(\Sigma^P)^{1/2}\dot{\bm s}_{(-1)} A^{+}\dot{\bm s}_{(-1)}^T(\Sigma^P)^{1/2}$ is symmetric and idempotent.  $\square$

\subsection{Further simulation studies}

In this subsection, we demonstrate the power of our transfer learning algorithms (cr-TLL) under the case when the regression coefficients of interest in the source model are similar to those in the target model. We also take the linear model as an example and compare our method with the maximum likelihood estimation (MLE) and the debiased method (DME) (see Section 5.1 for details).

The source model $P$ and target model $Q$ are set respectively as
\begin{align*}
  \text{Model }P: ~~&Y^P=\bm\beta^T X^P+\varepsilon^P,\\
  \text{Model }Q: ~~&Y^Q=\bm\gamma^T X^Q+\bm\theta^T Z^Q+\varepsilon^Q
\end{align*}
with $\varepsilon^P\sim N(0,1)$ and $\varepsilon^Q\sim N(0,1)$. Here the covariates $X^P=(X^{1P}, X^{2P}, X^{3P})^T\sim N(\bm{0},\Sigma_3)$ and $((X^Q)^T,(Z^Q)^T)=(X^{1Q}, X^{2Q}, X^{3Q}, Z^{1Q}, Z^{2Q})^T\sim N(\bm{0},\Sigma_5)$, where $\Sigma_d=(\sigma_{ij})_{1\leq i,j\leq d}$ with $\sigma_{ij}=0.2^{|i-j|}$. Moreover, the model parameters are set as $\bm\beta=(1,1,1)^T$, $\bm\gamma=(1,0.8,0.9)^T$ and $\bm\theta=(1,-1)^T$. Unlike the example in section 5.1.1, in the models above, the regression coefficients $\bm\beta$ and $\bm\gamma$ are similar.

The correlation-ratio matrix $¦«$ is estimated by the historical data of the variables $(X^P,Y^P)$ and  $(X^Q,Y^Q)$. The simulation results are reported in Table \ref{tab:9}. Similar conclusions as in section 5.1.1 could be drawn. In short, our method is even better than that designed specifically for similar models.

\begin{sidewaystable}[thp]  \scriptsize
\caption{\label{tab:9} The mean, sd and mse of the cr-TLL estimator, DME and MLE under the linear models when $\bm\beta$ and $\bm\gamma$ are similar.}
\centering
\begin{tabular}{|c|c|ccc|ccc|ccc|ccc|ccc| }
\hline
\multicolumn{17}{|c|}{Part I: fixed $n_Q=50$ and varying $n_P$}\\
\hline
\multicolumn{2}{|c|}{$n_P$}&\multicolumn{3}{|c|}{$50$}&\multicolumn{3}{|c|}{$100$}&\multicolumn{3}{|c|}{$200$}&
\multicolumn{3}{|c|}{$500$}&\multicolumn{3}{|c|}{$1000$}\\
\multicolumn{2}{|c|}{}& cr-TLL & DME&MLE & cr-TLL & DME&MLE & cr-TLL & DME&MLE & cr-TLL & DME&MLE & cr-TLL&DME&MLE \\
\hline
\multirow{5}{*}{mean}&$\gamma_1$&0.9766&1.0030&1.0014&0.9714&1.0021&1.0014&0.9712&1.0009&1.0014&0.9714&1.0014&1.0014&0.9704&1.0022&1.0014\\
&$\gamma_2$&0.8484&0.7990&0.7992&0.8571&0.7993&0.7992&0.8599&0.7997&0.7992&0.8605&0.8007&0.7992&0.8625&0.7998&0.7992\\
&$\gamma_3$&0.8638&0.8975&0.8968&0.8621&0.8968&0.8968&0.8600&0.8958&0.8968&0.8591&0.8959&0.8968&0.8587&0.8964&0.8968\\
&$\theta_1$&1.0220&1.0028&1.0028&1.0264&1.0047&1.0028&1.0329&1.0023&1.0028&1.0372&1.0047&1.0028&1.0384&1.0040&1.0028\\
&$\theta_2$&-0.9891&-0.9966&-0.9979&-0.9906&-0.9989&-0.9979&-0.9910&-0.9961&-0.9979&-0.9904&-0.9994&-0.9979&-0.9911&-0.9983&-0.9979\\
\hline
\multirow{5}{*}{sd}&$\gamma_1$&0.0952&0.1576&0.1534&0.0730&0.1556&0.1534&0.0544&0.1557&0.1534&0.0352&0.1565&0.1534&0.0256&0.1555&0.1534\\
&$\gamma_2$&0.0676&0.1565&0.1561&0.0494&0.1587&0.1561&0.0353&0.1576&0.1561&0.0221&0.1614&0.1561&0.0167&0.1578&0.1561\\
&$\gamma_3$&0.0562&0.1566&0.1551&0.0469&0.1567&0.1551&0.0366&0.1567&0.1551&0.0301&0.1577&0.1551&0.0283&0.1562&0.1551\\
&$\theta_1$&0.1017&0.1606&0.1593&0.0773&0.1610&0.1593&0.0615&0.1603&0.1593&0.0451&0.1674&0.1593&0.0372&0.1600&0.1593\\
&$\theta_2$&0.0496&0.1515&0.1515&0.0334&0.1518&0.1515&0.0245&0.1541&0.1515&0.0154&0.1554&0.1515&0.0111&0.1526&0.1515\\
\hline
\multirow{5}{*}{mse}&$\gamma_1$&0.0096&0.0248&0.0235&0.0061&0.0241&0.0235&0.0037&0.0242&0.0235&0.0020&0.0244&0.0235&0.0015&0.0241&0.0235\\
&$\gamma_2$&0.0069&0.0244&0.0243&0.0057&0.0251&0.0243&0.0048&0.0248&0.0243&0.0041&0.0260&0.0243&0.0041&0.0249&0.0243\\
&$\gamma_3$&0.0044&0.0245&0.0240&0.0036&0.0245&0.0240&0.0029&0.0245&0.0240&0.0025&0.0248&0.0240&0.0025&0.0244&0.0240\\
&$\theta_1$&0.0108&0.0257&0.0253&0.0066&0.0259&0.0253&0.0048&0.0256&0.0253&0.0034&0.0280&0.0253&0.0028&0.0256&0.0253\\
&$\theta_2$&0.0025&0.0229&0.0229&0.0012&0.0230&0.0229&0.0006&0.0237&0.0229&0.0003&0.0241&0.0229&0.0002&0.0232&0.0229\\
\hline
\hline
\multicolumn{17}{|c|}{Part II: fixed $n_P=1000$ and varying $n_Q$}\\
\hline
\multicolumn{2}{|c|}{$n_Q$}&\multicolumn{3}{|c|}{$10$}&\multicolumn{3}{|c|}{$50$}&\multicolumn{3}{|c|}{$100$}&
\multicolumn{3}{|c|}{$200$}&\multicolumn{3}{|c|}{$500$}\\
\multicolumn{2}{|c|}{}& cr-TLL & DME&MLE & cr-TLL & DME&MLE & cr-TLL & DME&MLE & cr-TLL & DME&MLE & cr-TLL&DME&MLE \\
\hline
\multirow{5}{*}{mean}&$\gamma_1$&0.9676&0.9931&0.9897&0.9686&1.0039&1.0042&0.9692&1.0001&1.0001&0.9698&0.9964&0.9972&0.9727&0.9989&0.9988\\
&$\gamma_2$&0.8630&0.8232&0.8234&0.8621&0.7961&0.7952&0.8612&0.7992&0.7996&0.8600&0.8038&0.8014&0.8560&0.7997&0.7998\\
&$\gamma_3$&0.8604&0.9147&0.9137&0.8587&0.8986&0.8984&0.8591&0.8959&0.8961&0.8594&0.9025&0.9013&0.8624&0.9030&0.9023\\
&$\theta_1$&1.0370&0.9872&0.9840&1.0359&1.0033&1.0029&1.0352&1.0024&1.0022&1.0327&1.0066&1.0043&1.0265&1.0001&0.9994\\
&$\theta_2$&-0.9899&-0.9815&-0.9827&-0.9903&-0.9938&-0.9951&-0.9902&-0.9998&-1.0003&-0.9901&-0.9976&-0.9989&-0.9899&-0.9993&-0.9994\\
\hline
\multirow{5}{*}{sd}&$\gamma_1$&0.0309&0.5064&0.5004&0.0259&0.1548&0.1541&0.0251&0.1085&0.1063&0.0244&0.0842&0.0735&0.0217&0.0473&0.0451\\
&$\gamma_2$&0.0173&0.5325&0.5292&0.0158&0.1628&0.1603&0.0155&0.1118&0.1092&0.0152&0.0962&0.0763&0.0149&0.0495&0.0482\\
&$\gamma_3$&0.0644&0.5354&0.5341&0.0279&0.1615&0.1618&0.0212&0.1090&0.1070&0.0168&0.0835&0.0760&0.0135&0.0477&0.0467\\
&$\theta_1$&0.0778&0.5197&0.5172&0.0384&0.1619&0.1587&0.0321&0.1107&0.1106&0.0285&0.0829&0.0743&0.0246&0.0476&0.0458\\
&$\theta_2$&0.0140&0.5238&0.5200&0.0113&0.1511&0.1502&0.0111&0.1065&0.1052&0.0108&0.0887&0.0729&0.0104&0.0486&0.0448\\
\hline
\multirow{5}{*}{mse}&$\gamma_1$&0.0020&0.2562&0.2503&0.0016&0.0239&0.0237&0.0015&0.0117&0.0113&0.0015&0.0071&0.0054&0.0012&0.0022&0.0020\\
&$\gamma_2$&0.0042&0.2838&0.2804&0.0041&0.0265&0.0257&0.0039&0.0124&0.0119&0.0038&0.0092&0.0058&0.0033&0.0024&0.0023\\
&$\gamma_3$&0.0057&0.2866&0.2852&0.0024&0.0260&0.0261&0.0021&0.0118&0.0114&0.0019&0.0069&0.0057&0.0015&0.0022&0.0021\\
&$\theta_1$&0.0074&0.2700&0.2675&0.0027&0.0262&0.0251&0.0022&0.0122&0.0122&0.0018&0.0069&0.0055&0.0013&0.0022&0.0021\\
&$\theta_2$&0.0002&0.2745&0.2704&0.0002&0.0228&0.0225&0.0002&0.0113&0.0110&0.0002&0.0078&0.0053&0.0002&0.0023&0.0020\\
\hline
\end{tabular}
\end{sidewaystable}

\end{document}